\documentclass[dvipsnames]{article}

\PassOptionsToPackage{numbers, compress}{natbib}
\usepackage[preprint]{neurips_2023}

\usepackage[utf8]{inputenc} 
\usepackage[T1]{fontenc}    
\usepackage{hyperref}       
\usepackage{url}            
\usepackage{booktabs}       
\usepackage{amsfonts}       
\usepackage{nicefrac}       
\usepackage{microtype}      
\usepackage{xcolor}         

\usepackage{graphicx}
\usepackage{xfrac}
\usepackage[autostyle]{csquotes}
\usepackage{enumitem}
\usepackage{hhline}
\usepackage{caption}
\usepackage{multirow}
\usepackage{multicol}
\usepackage{amssymb}
\usepackage{microtype}      
\usepackage{amsfonts}
\usepackage{caption}
\usepackage{subcaption}
\usepackage{wrapfig}
\usepackage{makecell}

\usepackage{chngcntr}

\usepackage{booktabs}
\usepackage[bottom]{footmisc}

\usepackage{amsmath}
\usepackage{amssymb}
\usepackage{mathtools}
\usepackage{amsthm}
\usepackage{bbm}

\usepackage[capitalize,noabbrev]{cleveref}

\usepackage{pifont}

\font\btt=rm-lmtk10

\newcommand{\mlp}{{\btt MLP}}

\theoremstyle{plain}

\theoremstyle{definition}

\theoremstyle{remark}

\usepackage{MnSymbol}%
\usepackage{wasysym}%
\usepackage[cal=dutchcal,
 calscaled=1,
 scr=euler]{mathalfa}
 
\usepackage{dsfont}

\newcommand{\mat}[1]{\ensuremath{{\mathbf{\MakeUppercase{{#1}}}}}}
\renewcommand{\vec}[1]{\ensuremath{\mathbf{\MakeLowercase{{#1}}}}}

\newcommand{\Wm}{\mat{W}}

\def\gC{{\mathcal{C}}}

\def\gF{{\mathcal{F}}}

\def\gL{{\mathcal{L}}}

\def\sC{{\mathbb{C}}}

\def\sR{{\mathbb{R}}}

\newcommand{\xv}{\vec{x}}

\newcommand{\cv}{\vec{c}}

\newcommand{\Nin}{\mathrm{N_{in}}}
\newcommand{\Nout}{\mathrm{N_{out}}}

\newcommand{\Lt}{\mathrm{L}}
\newcommand{\Nt}{\mathrm{N}}
\newcommand{\Dt}{\mathrm{D}}

\usepackage[textsize=tiny]{todonotes}

\usepackage{amsmath,amsfonts,amssymb, amsthm}
\usepackage{mathtools}

\hypersetup{colorlinks,citecolor={MidnightBlue},urlcolor={MidnightBlue}, linkcolor={black}}

\title{DNArch: Learning Convolutional Neural Architectures by Backpropagation}

\author{%
  David W.~Romero\thanks{Work done while interning at Google Research.} \\
  Vrije Universiteit Amsterdam \\
  Amsterdam, The Netherlands \\
  \texttt{d.w.romeroguzman@vu.nl} \\
  \And
  Neil Zeghidour \\
  Google Research \\
  Paris, France \\
  \texttt{neilz@google.com} \\
}

\begin{document}

\maketitle

\begin{abstract}
  We present \textit{Differentiable Neural Architectures} (DNArch), a method that jointly learns the weights and the architecture of Convolutional Neural Networks (CNNs) by backpropagation. In particular, DNArch allows learning (\textit{i}) the size of convolutional kernels at each layer,  (\textit{ii}) the number of channels at each layer, (\textit{iii}) the position and values of downsampling layers, and (\textit{iv}) the depth of the network. To this end, DNArch views neural architectures as continuous multidimensional entities, and uses learnable differentiable masks along each dimension to control their size. Unlike existing methods, DNArch is not limited to a predefined set of possible neural components, but instead it is able to discover entire CNN architectures across all feasible combinations of kernel sizes, widths, depths and downsampling. Empirically, DNArch finds performant CNN architectures for several classification and dense prediction tasks on sequential and image data. When combined with a loss term that controls the network complexity, DNArch constrains its search to architectures that respect a predefined computational budget during training.
\end{abstract}
\vspace{-3mm}
\section{Introduction}\label{sec:intro}
\vspace{-2mm}
Convolutional Neural Networks (CNNs) \cite{lecun1998gradient} are widely used for tasks such as image classification \cite{krizhevsky2012imagenet, he2016deep}, speech recognition \cite{sercu2016very, zeghidour2018fully}, text classification \cite{conneau2016very} and generative modeling \cite{oord2016wavenet, dhariwal2021diffusion} due to their performance and efficiency. However, tailoring a CNN architecture to a specific task or dataset typically requires substantial human intervention and cross-validation to design the architectures, e.g. to determine appropriate kernel sizes, width, depth, etc. This has motivated exploring the space of architectures in an automatic fashion, by developing architecture search algorithms~\citep{zoph17, liu18, bayesian_nas}.

\begin{figure}
\centering
\begin{minipage}{.55\textwidth}
  \centering
  \includegraphics[page=4, trim={0.5cm 21cm 16.5cm 0.5cm}, clip, width=\textwidth]{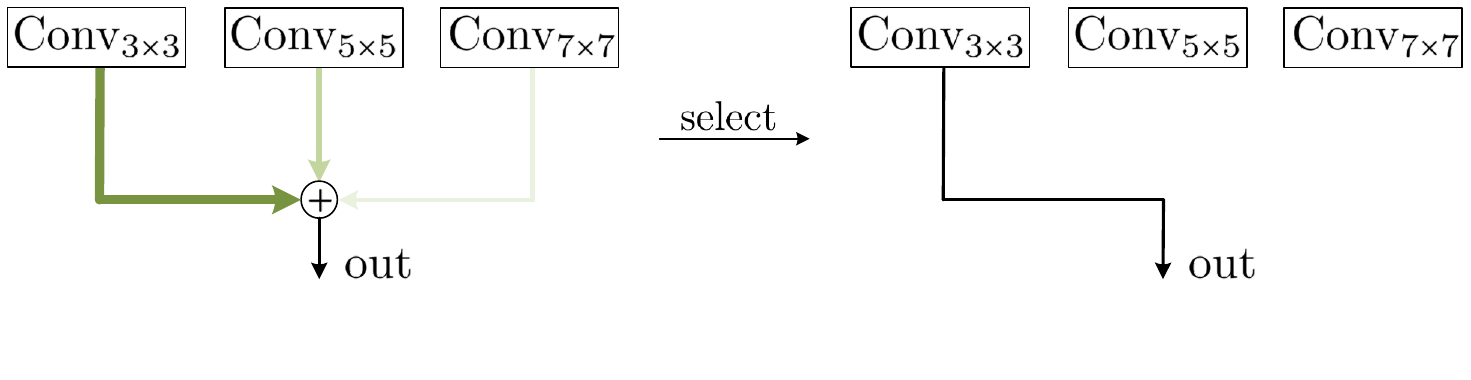}
  \vspace{-6.5mm}
    \captionof{figure}{DARTS learns the size of convolutional kernels using backpropagation to select among predefined options, e.g., DASH \cite{shen2022efficient}.}
    \label{fig:darts}
\end{minipage}\hfill
\begin{minipage}{.42\textwidth}
  \centering
  \hspace{5mm}
  \begin{subfigure}[b]{0.32\textwidth}
         \centering
         \includegraphics[page=2, trim={0.5cm 19cm 13cm 0.5cm}, clip, width=\textwidth]{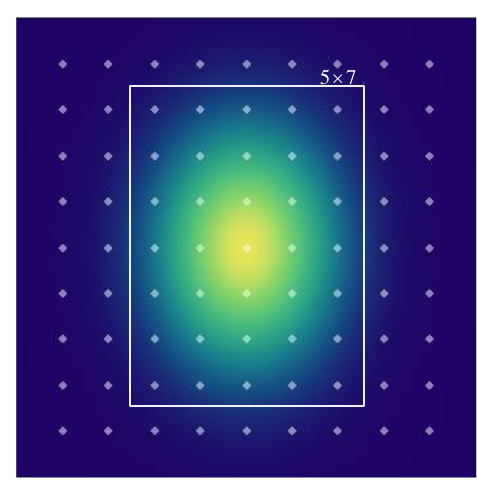}
     \end{subfigure}
     \hfill
     \begin{subfigure}[b]{0.32\textwidth}
         \centering
         \includegraphics[page=3, trim={0.5cm 19cm 13cm 0.5cm}, clip, width=\textwidth]{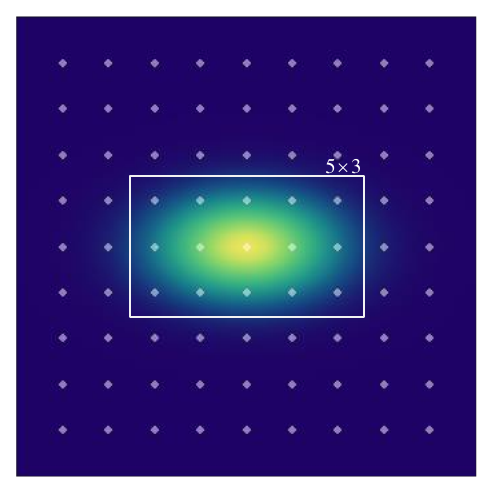}
     \end{subfigure}
     \hspace{5mm}
     \vspace{-2mm}
    \captionof{figure}{DNArch learns the size of convolutional kernels by modifying the parameters of the differentiable mask $m(\cdot \ ; \boldsymbol{\theta})$. Different $\boldsymbol{\theta}$ values lead to different sizes.}
    \label{fig:dnarch_kernels}
\end{minipage}
\vspace{-4mm}
\end{figure}

While these methods can find good architectures, they must solve an expensive discrete optimization problem that involves training and evaluating candidate architectures in each iteration, e.g. to optimize a reward with reinforcement learning \citep{zoph17}, or to evolve the model through a genetic algorithm \citep{liu18}. Differentiable Architecture Search (DARTS) \cite{liu2018darts} addresses this issue by allowing the network to consider a set of \textit{predefined} possible components in parallel, e.g., convolutions with kernels of size $3{\times}3$, $5{\times}5$, $7{\times}7$, and adjusting their contribution using learnable weights (Fig.~\ref{fig:darts}). Although DARTS is able to \textit{select} components via backpropagation, it requires (\textit{i}) defining a (small) set of possible components beforehand, (\textit{ii}) computing and keeping their responses in memory during training, and (\textit{iii})\break retraining the found architecture from scratch to remove the effect of other components in the output.
\begin{figure}
    \centering
    \includegraphics[page=1, trim={0.5cm 15cm 12cm 0.5cm}, clip, width=0.95\textwidth]{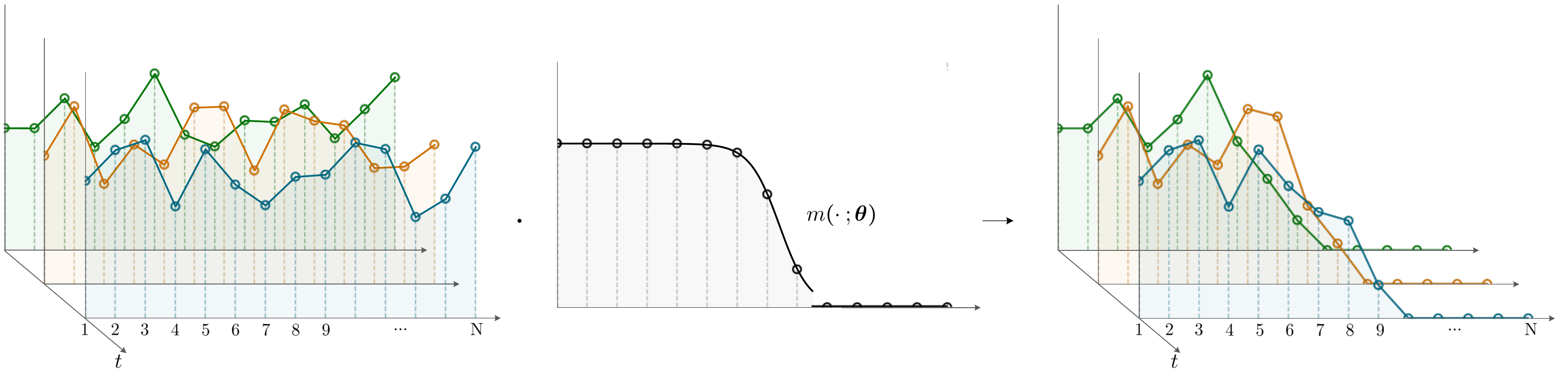}
    \vspace{-4mm}
    \caption{DNArch views neural architectures as entities in a continuous multidimensional space, and uses differentiable masks to learn their length by backpropagation. In this example, DNArch learns the width of a layer by applying a differentiable mask $m$ with learnable parameters $\boldsymbol{\theta}$ to the channel dimension. As a result, changes in the value of $\boldsymbol{\theta}$ effectively results in changes in the layer's width.
    \vspace{-4mm}
    }\label{fig:figure_1}
\end{figure}

In this paper, we introduce \textit{Differentiable Neural Architectures} (DNArch), a method that simultaneously learns the weights and the entire architecture of a CNN during training by backpropagation. Specifically, DNArch learns the weights as well as (\textit{i}) the size of convolutional kernels at each layer, (\textit{ii}) the number of channels at each layer, (\textit{iii}) the position and resolution of downsampling layers, and (iv) the number of layers of the network. 
To this end, DNArch takes a novel approach to learning neural architectures by viewing them as entities defined in a multidimensional continuous space with dimensions corresponding to network attributes, e.g., depth, width, etc., and using differentiable masks with learnable parameters along each dimension to control their length (Fig.~\ref{fig:figure_1}). Unlike DARTS methods, e.g., \cite{liu2018darts, shen2022efficient}, DNArch does \textit{not} require a predefined set of components to choose from, but instead is able to explore among \textit{all} feasible values, e.g., all kernel sizes between $1{\times}1$ and $\mathrm{N}{\times}\mathrm{N}$ for a $\mathrm{N}{\times}\mathrm{N}$ image. This is a result of the truly continuous nature of DNArch, which, unlike DARTS, does not require multiple instantiations of the same layer for different parameter values (Fig.~\ref{fig:darts}). Instead, DNArch explores the parameter space by modifying the learnable parameters of the differentiable masks (Fig.~\ref{fig:dnarch_kernels}), making it a much more scalable NAS method. Moreover, since both the architecture and the weights are optimized in a single run, no retraining is needed after training.


\textbf{Results.} We empirically show that DNArch is able to find performant CNN architectures across several classification and dense prediction tasks on sequential and image datasets. The architectures found by DNArch consistently surpass the general-purpose convolutional architecture on which DNArch is applied, and often outperform specialized task-specific architectures. Moreover, we show that DNArch can be easily combined with a regularization term that controls the computational complexity of candidate networks. By doing so, DNArch explores among neural architectures that respect a predefined computational budget during the entire training process. As a result, finding architectures with DNArch is roughly as expensive as a single training loop of the underlying baseline.


\vspace{-3mm}
\section{Method}
\vspace{-2mm}
DNArch has two key components: Differentiable Masking and Continuous Kernel Convolutions. Here, we introduce these concepts and show how they can be used to learn CNN architectures next.
\vspace{-3mm}
\subsection{Differentiable Masking}\label{sec:diff_masking}
\vspace{-2mm}
Let us consider an arbitrary function $f: [\mathrm{a}, \mathrm{b}] \rightarrow \sR$, which we want to be non-zero only in a subset $[\mathrm{c}, \mathrm{d}] \subseteq [\mathrm{a}, \mathrm{b}]$. To this end, we can multiply $f$ with a mask $m$ whose values are non-zero only on $[\mathrm{c}, \mathrm{d}]$, e.g., a rectangular mask $\Pi_{[\mathrm{c}, \mathrm{d}]}(x) {=} \mathbbm{1}_{[\mathrm{c}, \mathrm{d}]}$. However, as its gradient is either zero or non-defined, it is not possible to learn the interval in which it is non-zero by backpropagation.
To overcome this limitation, we can instead use a parametric differentiable mask $m(\cdot\ ; \boldsymbol{\theta})$ whose interval of non-zero values is defined by its parameters $\boldsymbol{\theta}$. As the mask $m(\cdot\ ; \boldsymbol{\theta})$ is differentiable with regard to its parameters $\boldsymbol{\theta}$, we can learn the interval on which it is non-zero using backpropagation.

In this work, we consider two types of  masks: a Gaussian mask $m_{\mathrm{gauss}}\left(\cdot \ ; \{\mu, \sigma^2\}\right)$ parameterized by its mean and variance $\boldsymbol{\theta}{=}\{\mu, \sigma^2\}$; and a Sigmoid mask $m_\mathrm{sigm}\left(\cdot \ ;\{\mu, \tau\}\right)$ parameterized by its offset and its temperature $\boldsymbol{\theta}{=}\{\mu, \tau\}$ defined as:
\begin{align}
 m_{\mathrm{gauss}}(x \ ; \{\mu, \sigma^2\})&=\left\{ \exp\hspace{-0.5mm}\left( - \tfrac{1}{2} \tfrac{(x - \mu)^2}{\sigma^2}\right) \ \ \text{if}\  \exp\hspace{-0.5mm}\left( - \tfrac{1}{2} \tfrac{(x - \mu)^2}{\sigma^2}\right) \geq T_m; \  0 \  \text{otherwise} \right\}, \label{eq:gauss_mask}\\
m_\mathrm{sigm}\left(x \ ;\{\mu, \tau\}\right)&=\left\{1 - \mathrm{sigm}\left(\tau (x - \mu) \right) \ \ \text{if}\ 1 - \mathrm{sigm}\left(\tau (x - \mu) \right)  \geq T_m; \  0 \  \text{otherwise} \right\}, \label{eq:sigm_mask}
\end{align}

\begin{wrapfigure}{r}{0.45 \textwidth}
    \centering
    \begin{subfigure}[b]{0.23\textwidth}
         \centering
         \includegraphics[page=5, trim={0.5cm 19cm 6.2cm 2cm}, clip, width=\textwidth]{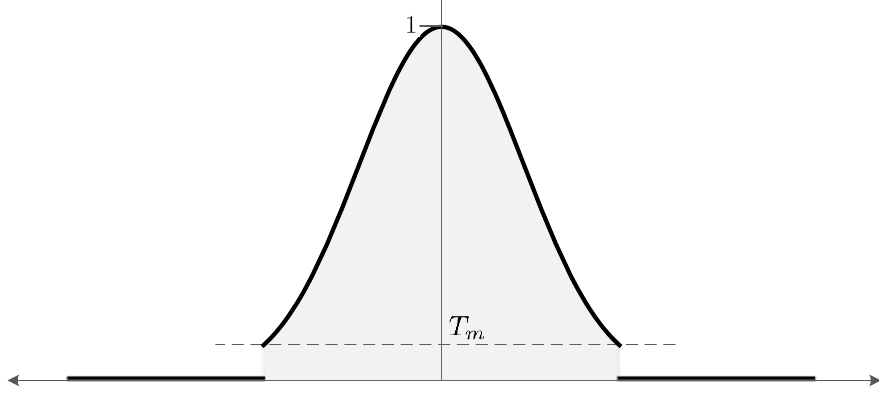}
     \end{subfigure}
     \hfill
     \begin{subfigure}[b]{0.20\textwidth}
         \centering
         \includegraphics[page=6, trim={1.2cm 19cm 6.2cm 2cm}, clip, width=\textwidth]{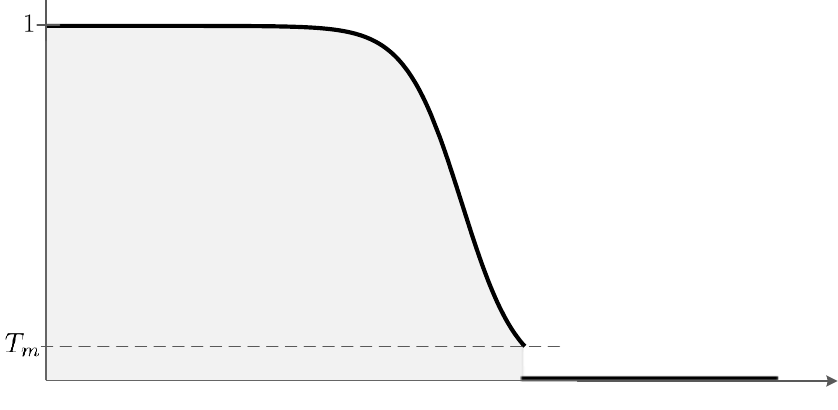}
     \end{subfigure}
     \vspace{-4mm}
     \caption{Gaussian and sigmoid masks.
     \vspace{-5mm}}
    \label{fig:diff_masks}
\end{wrapfigure}
where $T_m$ is a predefined threshold below which the mask is zero. These masks are illustrated in Fig.~\ref{fig:diff_masks}. To avoid clutter, in the rest of the document we will refer to these masks as $m_{\mathrm{gauss}}$ and $m_{\mathrm{sigm}}$, and will provide specific instantiations when needed.

\textbf{Multidimensional masks.} $\mathrm{N}$-dimensional masks can be constructed by combining $\mathrm{N}$ 1D masks, each with their own parameters. For example, the Gaussian mask used to learn the size of convolutional kernels in Fig.~\ref{fig:dnarch_kernels} is constructed as:
\begin{equation}
\setlength{\abovedisplayskip}{4pt}
\setlength{\belowdisplayskip}{4pt}
m_\mathrm{gauss}(x, y; \left\{\{\mu_X, \mu_Y\}, \{\sigma^2_X, \sigma^2_Y\} \right\}) = m_{\mathrm{gauss}}\left(x \ ; \{\mu_X, \sigma^2_X\}\right) \cdot m_{\mathrm{gauss}}\left(y \ ; \{\mu_Y, \sigma^2_Y\}\right)
\end{equation}

\vspace{-3mm}
\subsubsection{Materializing parameters only for non-zero mask values}\label{sec:only_nonzero_mask}
\vspace{-1mm}
Parts of differentiable masks will map to zero based on the value of the parameters $\boldsymbol{\theta}$. Therefore, it would be a waste of compute and memory to materialize the mask --and the corresponding network parameters, e.g., channels $\mathrm{ch} \in [\mathrm{10}, \mathrm{N}]$ in Fig.~\ref{fig:figure_1}-- to zero them out next. Luckily, we can take advantage of the invertible form of the Gaussian and Sigmoid masks to materialize parameters only for values for which the mask is non-zero. To this end, we find the value $x_{T_m}$ for which the mask is equal to the threshold $T_m$, i.e., $x_{T_m} {=} x \ \text{such that}\ m(x;  \boldsymbol{\theta}){=} T_m$, and only materialize the mask and the corresponding network parameters for values of $x$ for which the value of the mask is greater than $T_m$. By inverting the mask equations (Eqs.~\ref{eq:gauss_mask},~\ref{eq:sigm_mask}),~we~obtain~$x_{T_m}$~as:
\begin{center}
\vspace{-7mm}
    \begin{tabular}{p{5cm}p{6cm}}
    \begin{equation}
        \pm x_{T_m} = \mu \pm \sqrt{-2 \sigma^2 \log(T_m)}, \label{eq:inv_gauss_mask}
    \end{equation}
    &
    \vspace{-1mm}
    \begin{equation}
         x_{T_m} = \mu - \tfrac{1}{\tau}\log\left( \tfrac{1}{1 - T_m} - 1\right),\label{eq:inv_sigm_mask}
    \end{equation}
    \end{tabular}
    \vspace{-8mm}
\end{center}
for Gaussian and Sigmoid masks, respectively. Consequently, we can make sure that all rendered values will be used by only materializing the mask and related network parameters for values of $x$ within the range $[-x_{T_m}, x_{T_m}]$ for Gaussian masks and $[x_\mathrm{min}, x_{T_m}]$ for Sigmoid masks, where $x_\mathrm{min}$ depicts the lowest coordinate indexing the mask.
\vspace{-3mm}
\subsection{Continuous Kernel Convolutions}
\vspace{-2mm}
To prevent finding poor architectures due to insufficiently large receptive fields, it is important for a network to be able to model the global context regardless of specific architectural choices. We rely on Continuous Kernel Convolutions (CKConvs) \cite{romero2021ckconv} to model global dependencies on inputs of arbitrary length, resolution and dimensionality regardless of the network architecture \cite{knigge2023modelling}. CKConvs view convolutional kernels $\boldsymbol{\psi}$ as continuous functions parameterized by a small neural network $\text{\mlp}_{\psi}: \sR^{\Dt} \rightarrow \sR^{\Nin \times \Nout}$ that receives coordinates $\cv_i \in \sR^{\Dt}$ as input and predicts the value of the convolutional kernel at those coordinates: $\cv_i \mapsto \text{\mlp}_{\psi}(\cv_i){=}\psi(\cv_i)$. To construct a kernel of size $\mathrm{K_x}{\times}\mathrm{K_y}$, a CKConv layer constructs a grid of $\mathrm{K_x}{\times}\mathrm{K_y}$ coordinates $[\cv_{(1,1)}, \cv_{(1,2)}, ..., \cv_{(\mathrm{K_x}, \mathrm{K_y})}]$, and passes each coordinate through the neural network $\text{\mlp}_{\psi}$ (Fig.~\ref{fig:learnable_kernel_sizes}). As a result, CKConvs construct large kernels with few parameters by detaching the size of the kernel from its parameter count.
\vspace{-3mm}
\subsection{The need for learnable architectures}\label{sec:need_for_learnable_archs}
\vspace{-1mm}
General-purpose architectures like Perceiver \cite{jaegle2021perceiver} and the Continuous CNN (CCNN) \cite{romero2022towards} make few assumptions about their input signals, and thus require few architectural changes to handle different tasks. However, the architectures of general-purpose models are static, and thus they are likely not optimal among all the tasks the model might need to solve.
For instance, Perceiver maps inputs to a hidden representation of constant size regardless of the complexity of the task and the input length, resolution and dimensionality. Consequently, it will likely not be able to represent tasks on large inputs correctly. CCNNs, on the other hand, avoid pooling and always perform (global) convolutions on the original input resolution. While this addresses the issue of having a hidden representation of fixed size for inputs with different characteristics, CCNNs can lead to unnecessarily high computational complexity by always performing convolutions at the input resolution. In addition, both the architectures of Perceivers and CCNNs are controlled by non-differentiable hyperparameters, and thus adapting them to a new task requires hyperparameter search across many configurations. 
To address these limitations, we instead propose to construct neural architectures able to tune themselves to fit the requirements of a particular task in an efficient manner, i.e., in a single training run.
\vspace{-3mm}
\subsection{Learning CNN architectures by backpropagation}
\vspace{-1mm}
Most components of DNArch, such as the learning of the network's width and depth, are not limited to convolutional architectures. However, this research aims to show \textit{how DNArch can be used to learn as many components of a neural architecture as possible}. To that end, we use a general-purpose convolutional architecture: the CCNN \cite{knigge2023modelling}, and make all its architectural components learnable.

\begin{wrapfigure}{r}{0.42\textwidth}
\vspace{-5mm}
    \centering
    \includegraphics[page=8, trim={0.5cm 13cm 3.8cm 0.5cm}, clip, width=0.39\textwidth]{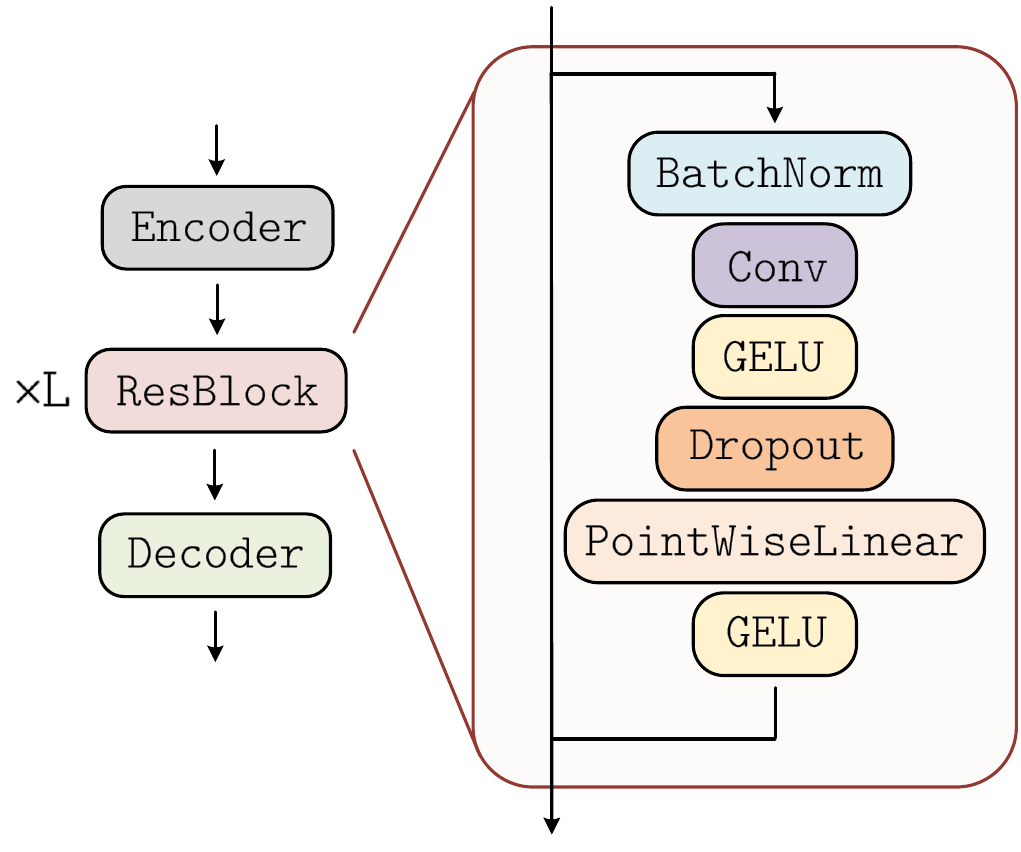}
    \vspace{-3mm}
    \caption{The CCNN architecture \citep{romero2022towards}.
    \vspace{-4mm}}
    \label{fig:ccnn_architecture}
\end{wrapfigure}

\textbf{The Continuous CNN \cite{knigge2023modelling}.} The Continuous CNN (CCNN) is a general-purpose convolutional model able to handle inputs of arbitrary dimension, length and resolution without changes. It consists of an {\tt Encoder}, a {\tt Decoder}, and many residual blocks (Fig.~\ref{fig:ccnn_architecture}). We refer to the branch on which residual blocks modify the input as \textit{residual branch}, and to the branch connecting the input and the output directly as the \textit{identity branch}. The {\tt Encoder} and {\tt Decoder} adapt the input and output of the model to the goal of the task, e.g., dense / global predictions. Importantly, CCNN's ability to model global context on inputs of any resolution, length and dimensionality makes it an ideal base network for DNArch as (\textit{i}) it prevents the formation of poor architectures due to insufficient receptive fields, and (\textit{ii}) it allows DNArch to be used on tasks on data of arbitrary length and dimensionality without changing the base network --which is needed in existing methods (see NAS-Bench-360 \cite{tu2022bench} for several examples).
\vspace{-3mm}
\subsubsection{Learning the size of convolutional kernels}\label{sec:learn_kernel_sizes}
\vspace{-2mm}
\begin{wrapfigure}{r}{0.42 \textwidth}
\vspace{-6mm}
    \centering
    \includegraphics[page=14, trim={2.5cm 15.5cm 20cm 0.5cm}, clip, width=0.37\textwidth]{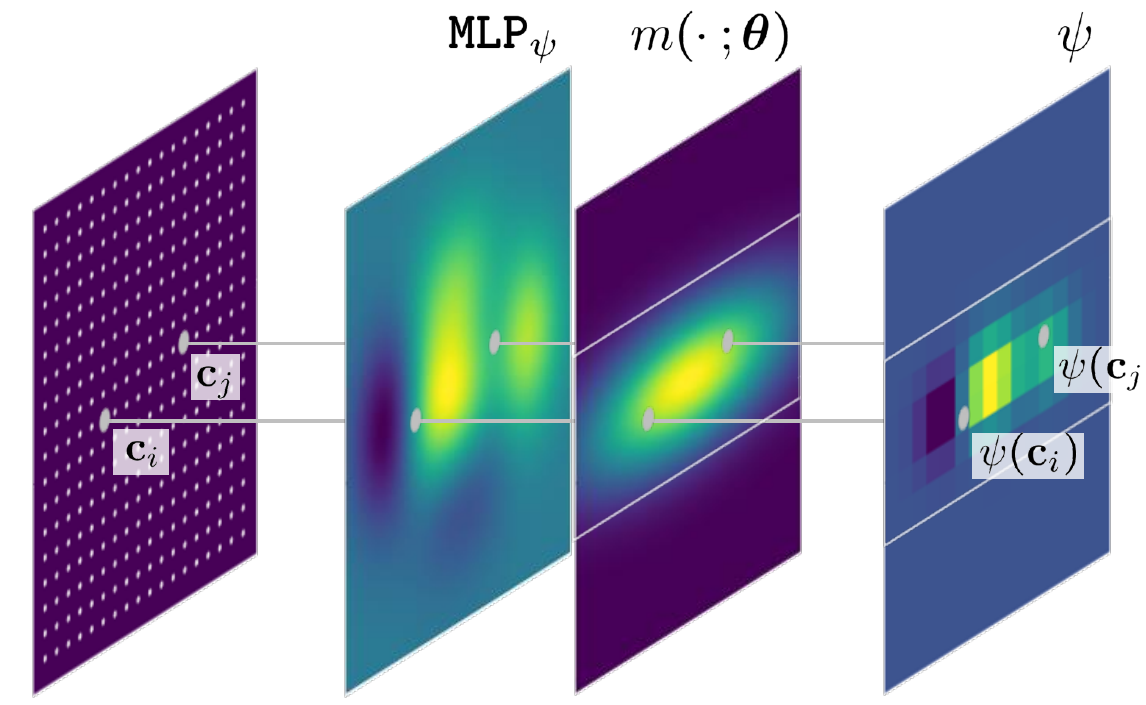}
    \vspace{-2.5mm}
    \caption{Learning kernel sizes with differentiable masking and CKConvs \citep{romero2021ckconv}.
    \vspace{-5mm}}
    \label{fig:learnable_kernel_sizes}
\end{wrapfigure}

First introduced in FlexConv \cite{romero2021flexconv}, differentiable masking can be combined with CKConvs to learn the size of convolutional kernels by backpropagation. This is done by modelling convolutional kernels $\psi$ as the product of a small neural network $\text{\mlp}_\psi$, i.e., a Continuous Kernel Convolution, and a differentiable mask $m(\cdot\ ;\boldsymbol{\theta})$ with learnable parameters, i.e., $\psi(\cv_i) {=} {\text{\mlp}}_\psi(\cv_i) \cdot m(\cv_i; \boldsymbol{\theta})$  (Fig.~\ref{fig:learnable_kernel_sizes}). Note that, it is possible to construct the convolutional kernel only for non-zero values of the mask $m(\cv_i; \boldsymbol{\theta})$ by following the method outlined in Sec.~\ref{sec:only_nonzero_mask}.

\vspace{-3mm}
\subsubsection{Learning downsampling layers}
\vspace{-2mm}
We can also use differentiable masking to learn downsampling by applying a differentiable mask on the Fourier domain. The Fourier transform $\gF$ represents a function $f: \sR^{\Dt} \rightarrow \sR$ in terms of its\break \textit{spectrum} $\tilde{f}: \sR^{\Dt} \rightarrow \sC$, which map frequencies $\omega$ to the amount of that frequency in the input $\tilde{f}(\omega)$. A useful identity in this context is that cropping high frequencies in the Fourier domain equals downsampling in the spatial domain.

\begin{wrapfigure}{r}{0.55\textwidth}
\vspace{-4mm}
    \centering
    \includegraphics[page=12, trim={0cm 5cm 15cm 0.5cm}, clip, width=0.55\textwidth]{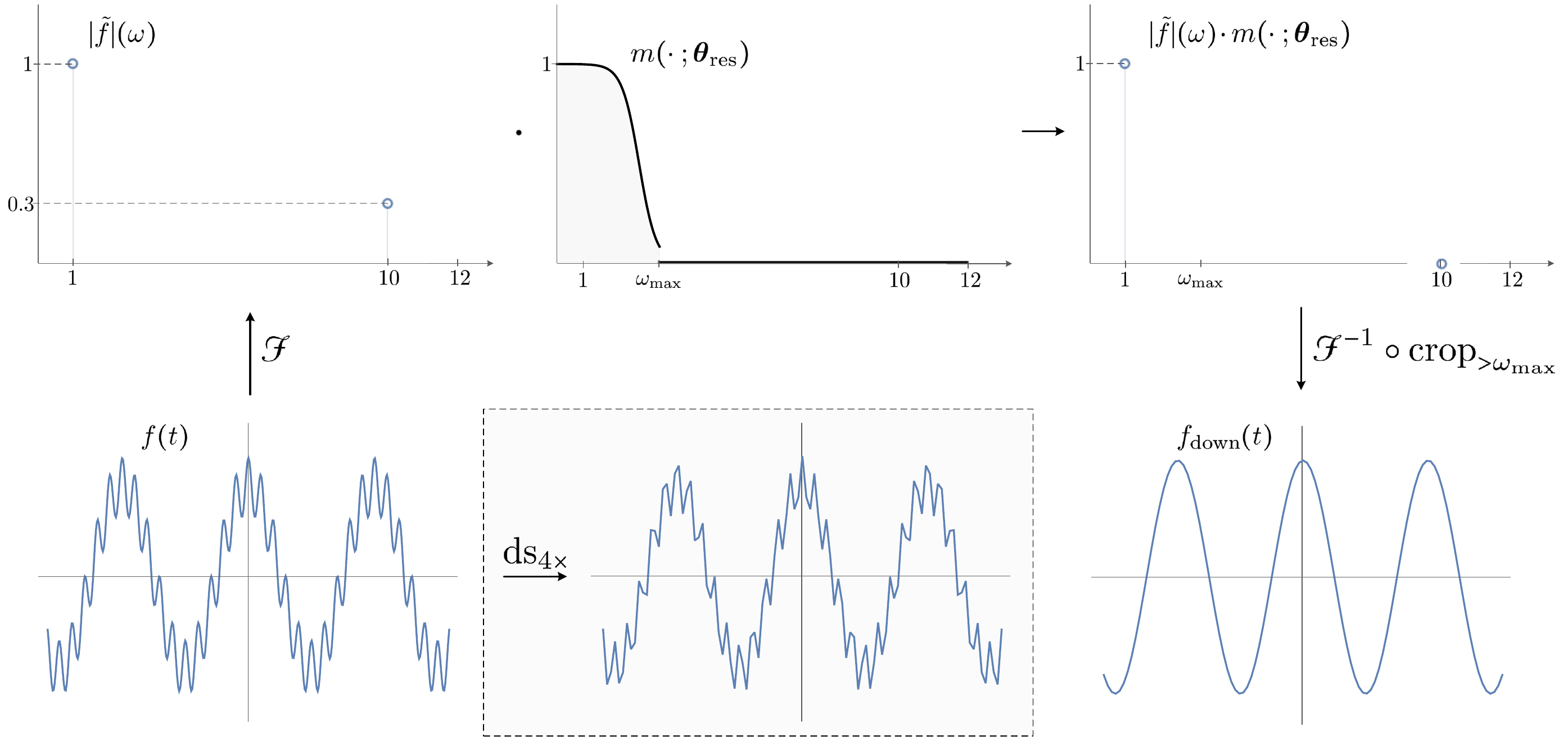}
    \vspace{-6mm}
    \caption{Learning downsampling with differentiable masking on the Fourier domain.
    \vspace{-5mm}}
    \label{fig:learnable_downsample}
\end{wrapfigure}
To learn downsampling, we use a learnable sigmoid mask $m_\mathrm{sigm}$ to perform a \emph{learnable low-pass filtering} on the input. This is achieved by multiplying the spectrum of the input with the mask $m_\mathrm{sigm}$. By doing so, all frequencies above the mask's cutoff frequency $\omega_\mathrm{max}{=}T_m$ becomes zero (Fig.~\ref{fig:learnable_downsample}). An important consequence of low-pass filtering is that as the spectrum of the signal becomes zero above $\omega_\mathrm{max}$, the low-passed signal can be faithfully represented at a lower resolution determined by $\omega_\mathrm{max}$. Letting $\gF$, $\gF^{-1}$ be the Fourier and inverse Fourier transform, $\mathrm{crop}_{> \omega_\mathrm{max}}$ be an operator that crops all values above $\omega_\mathrm{max}$, and $f_\mathrm{down}$ represent the downsampled signal $f$, we have that: 
\begin{equation}
\setlength{\abovedisplayskip}{4pt}
\setlength{\belowdisplayskip}{4pt}
f_\mathrm{down} {=} \gF^{-1}\left[ \mathrm{crop}_{>\omega_\mathrm{max}}\left(\gF[f] \cdot m_\mathrm{sigm}( \ \cdot ; \boldsymbol{\theta}) \right) \right].
\end{equation}
Fig.~\ref{fig:learnable_downsample} shows an example of downsampling by a factor of $4\mathrm{x}$. Unlike conventional downsampling, e.g., max-pooling, \textit{spectral downsampling} \cite{rippel2015spectral, riad2022learning} considers the spectral content of the input during downsampling, and thus prevents \textit{aliasing} --where the output resolution is insufficient to accurately represent the underlying signal (Fig.~\ref{fig:learnable_downsample}, middle down)--. This is important since it has been shown that aliasing has negative effects on robustness \cite{zhang2019making}, generation \cite{karras2021alias}  and generalization \cite{vasconcelos2021impact}.

\textbf{Combining learnable downsampling and convolution.} The previous method requires mapping inputs to the Fourier domain and back to learn downsampling. Fortunately, CCNNs as well as most methods that rely on global convolutions, e.g., CKConv \citep{romero2021ckconv}, S4 \citep{gu2021efficiently}, rely on the \textit{Fourier convolution theorem}: $(f * \psi) {=} \gF^{-1}\left[\gF[f] \cdot \gF[\psi] \right]$ to compute convolutions with large kernels efficiently. This means that CCNNs already use a Fourier and inverse Fourier transforms in each residual block to compute convolutions. Hence, we can avoid recomputing these steps by placing the learnable downsampling operation \textit{within the Fourier convolution}. Specifically, we can simultaneously compute downsampling and convolution by applying the differentiable mask $m_\mathrm{sigm}$ and the cropping operations $\mathrm{crop}_{>\omega_\mathrm{max}}$ before returning from the Fourier domain back to the spatial domain. That is:\footnote{We note that the Fourier transform is not strictly necessary to learn downsampling, e.g., for CNNs with local kernels. Leveraging the Fourier convolution theorem, equivalent downsampling can be achieved by convolving the input with the inverse Fourier transform of the mask in the spatial domain (see Appx.~\ref{appx:downsample_no_fourier} for details).}
\begin{equation}
\setlength{\abovedisplayskip}{4pt}
\setlength{\belowdisplayskip}{4pt}
    (f * \psi)_\mathrm{down} = \gF^{-1}\left[\mathrm{crop}_{>\omega_\mathrm{max}}(m_\mathrm{sigm}(\cdot\ ;\boldsymbol{\theta}) \cdot \gF[f] \cdot \gF[\psi]) \right].\label{eq:conv_and_downsampling}
\end{equation}
%
\textbf{Materializing functions only on the output resolution.} Note that Eq.~\ref{eq:conv_and_downsampling} computes the convolution on the resolution of the input and downsamples next. This incurs in an unnecessary overhead as the output of the convolution will be downsampled directly after. A more efficient approach comes from inverting the order of these operations to compute the convolution at the downsampled resolution. Luckily, this can be achieved by using the method outlined in Sec.~\ref{sec:only_nonzero_mask}. Since the cutoff frequency of the mask corresponds to the coordinate at which the mask equals the threshold, i.e., $\omega_\mathrm{max} {=} x_{T_m}$, it can be calculated using Eq.~\ref{eq:inv_sigm_mask}. Next, since the cutoff frequency defines the minimum resolution required to faithfully represent the input, we can simply downsample the input and convolutional kernel to that resolution before the convolution to compute it on the output resolution.

\textbf{Learning subsampling for dense tasks.}  \citet{riad2022learning} apply learned downsampling on both the identity and the residual branches of a residual block to limit resolution of all representations after a specific residual block. In the context of DNArch, this is undesirable for two reasons: First, as we learn the whole network architecture during training, it is not known a priori what resolution mappings will require at each layer. However, forcing the identity branch to have the same resolution as the corresponding residual branch restricts all subsequent mappings to be of maximum that resolution. Secondly, dense prediction tasks, e.g., segmentation, require the learned architecture to produce outputs that share the same resolution as the input. However, if the identity branch is also downsampled, the output of the network would be of lower resolution even for a single level of downsampling in the network. This in turn, would result in over-smoothed predictions.

Based on these observations, we use downsampling \textit{only} on the residual branch and upsample features at the end of each residual block back to the resolution of the input. This allows us to (\textit{i}) have features at same resolution as the input in the last layer, and (\textit{ii}) learn U-Net \cite{ronneberger2015u} like architectures. 

\vspace{-3mm}
\subsubsection{Learning the width of the network at each layer}
\vspace{-1mm}
We learn the width of a layer by applying a differentiable mask $m(\cdot\ ;\boldsymbol{\theta})$ along the channel dimension of feature representations (see Fig.~\ref{fig:figure_1}). The network can then adjust the width of its representations by changing the value of $\boldsymbol{\theta}$. By using the method in Sec.~\ref{sec:only_nonzero_mask} only non-masked channels are rendered.

\begin{wrapfigure}{r}{0.42\textwidth}
\vspace{-6mm}
    \centering
    \includegraphics[page=9, trim={13.8cm 2.5cm 16cm 0.5cm}, clip, width=0.39\textwidth]{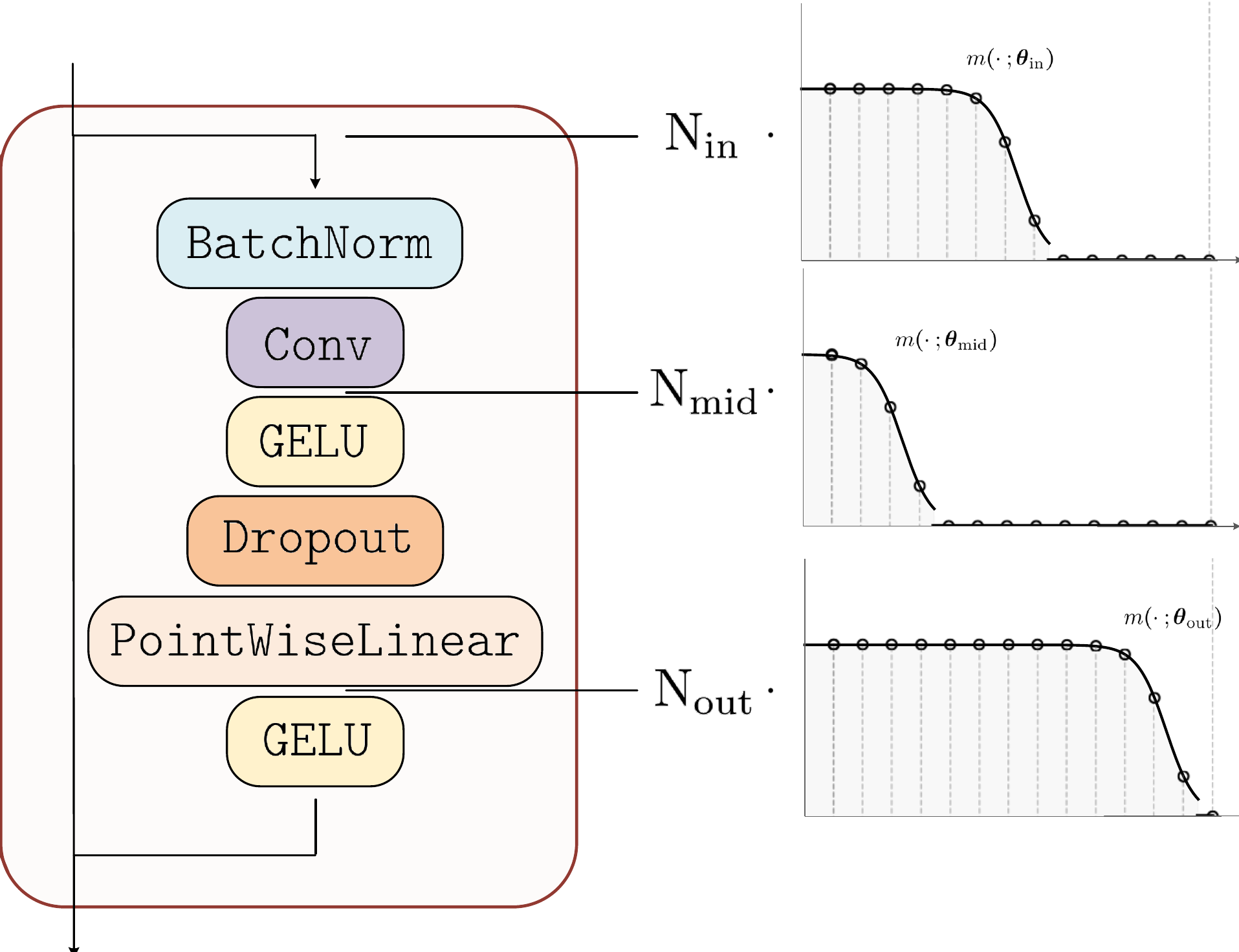}
    \vspace{-4mm}
    \caption{Positioning of the width masks. 
    \vspace{-6mm}}
    \label{fig:width_learning}
\end{wrapfigure}

\textbf{Positioning of the width masks.} We aim to learn the width of all layers in a network. To this end, we apply differentiable masks with independent learnable parameters along the channel dimensions of all the network components that change the network's width, i.e., all {\tt Conv} and {\tt PWLinear} layers. This corresponds to learning three independent masks for each residual block in the network, which correspond to the input ($\mathrm{N_{in}}$), the middle ($\mathrm{N_{mid}}$) and the output ($\mathrm{N_{out}}$) channels of the residual block (Fig.~\ref{fig:width_learning}). 
Components that do not change the width, e.g., {\tt BatchNorm}, {\tt GELU}, have their width determined by the preceding differentiable mask.

\textbf{The advantage of avoiding masks on the identity branch.} Applying a differentiable mask on the output of the entire residual block to constrain its width, i.e., after the sum of the residual and identity branches, would accumulate the effect of all masks applied before that block. As a result, the $l$-th block would be effectively masked by the combination of all masks before that block, i.e., $\prod_{i \leq l} m_i$, with $m_i$ the mask after the $i$-th block. Since the values of the masks live in the $[0, 1]$ interval, this would result in an exponential decrease in the magnitude of the hidden feature presentations. To avoid this, we apply the masks on the residual branch only. In addition, keeping the identity branch intact allows DNArch to construct DenseNet-like architectures \cite{huang2017densely}, where blocks can reuse channels that only have been modified by some --or none-- of the previous blocks.
\vspace{-3mm}
\subsubsection{Learning the depth of the network}\label{sec:learning_depth}
\vspace{-1mm}
\begin{wrapfigure}{r}{0.32 \textwidth}
\vspace{-6mm}
    \centering
    \includegraphics[page=10, trim={0.5cm 6cm 8.5cm 0.5cm}, clip, width=0.168\textwidth]{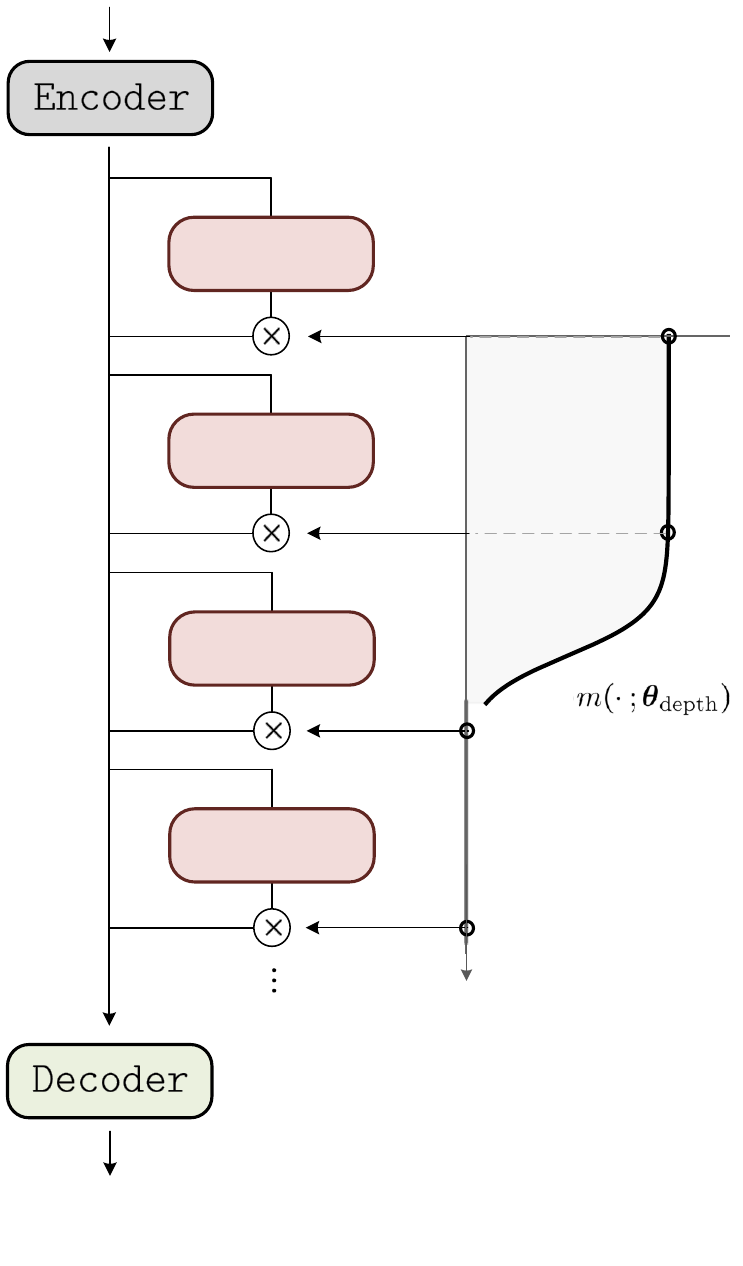}
    \vspace{-5mm}
    \caption{Learning network's depth with differential masking.
    \vspace{-12mm}}
    \label{fig:depth_learning}
\end{wrapfigure}

We learn the network's depth by viewing the number of residual blocks as a continuous axis with values $[1, 2, .., \Dt]$ corresponding to the index of each block, and using a differentiable mask $m(\cdot, \boldsymbol{\theta})$ along this axis to dynamically mask out blocks based on the value of the mask parameters $\boldsymbol{\theta}$ (Fig.~\ref{fig:depth_learning}).

\textbf{Positioning of the depth mask.} To ensure that information flows from the input to the output of the network regardless of the value of the mask parameters, we \textit{only} apply the mask on the residual branch. If the mask were also applied on the identity branch, feature representations at the end of the network could become zero, and the network would only be able to output random predictions.

\vspace{-3mm}
\subsubsection{Putting it all together: Learning entire convolutional architectures by backpropagation}
\vspace{-1mm}
By simultaneously using the methods outlined in Secs.~\ref{sec:learn_kernel_sizes}-\ref{sec:learning_depth}, DNArch uses backpropagation to learn the weights, the size of convolutional kernels at each layer, the number of channels at each layer, the position and resolution of downsampling layers, and the depth of a convolutional network.
\vspace{-3mm}
\subsection{Learning convolutional architectures under computational constraints}
\vspace{-1mm}
We can ensure that the architectures searched by DNArch respect a predefined computational complexity by including an additional regularization term $\gL_\mathrm{comp}$ that reflects the complexity of the current candidate architecture based on its mask parameters. To this end, we define the optimization loss $\gL$ as the sum of the task objective loss $\gL_\mathrm{obj}$ and the complexity loss $\gL_\mathrm{comp}$~weighted~by~a~factor~$\lambda$:
\begin{equation}
\setlength{\abovedisplayskip}{4pt}
\setlength{\belowdisplayskip}{4pt}
    \gL = \gL_\mathrm{obj} + \lambda \ \gL_\mathrm{comp}.\label{eq:optim_loss}
\end{equation}
By minimizing this loss, DNArch is encouraged to find architectures that meet the desired computational budget while still achieving good performance on the end task.

\vspace{-3mm}
\subsubsection{Defining the complexity loss $\gL_\mathrm{comp}$} 
\vspace{-1mm}
The purpose of $\gL_\mathrm{comp}$ is to use the size of the learned masks to estimate the total computation needed for a forward pass of the network. Its construction is outlined below.

\textbf{Layer-wise complexities.} Let $\gC_{\mathrm{layer}}(\mathrm{L}, \Nin, \Nout)$ be the number of operations required in a given layer with an input of length $\mathrm{L}$ and $\Nin$ and $\Nout$ input and output channels. To estimate the number of computations required based on the current size of the masks, we can substitute the lenght of each dimension with the size of the corresponding masks: $\gC_{\mathrm{layer}}\left(\mathrm{size}(m_\mathrm{res}), \mathrm{size}(m_\Nin), \mathrm{size}(m_\Nout)\right)$. As an example, consider a pointwise linear layer. A pointwise linear layer $\mathrm{lin}: \sR^{\Nin} \rightarrow \sR^{\Nout}$ takes an input $f$ of length $\Lt$ and $\Nin$ channels and multiplies each element along the spatial dimensions of the input with a matrix of dimensions $[\Nin, \Nout]$ to produce an output of the same length, but with $\Nout$ number of channels.
The total operations required in this layer is given by $
\gC_{\mathrm{lin}}(f) {=} \Lt \cdot \Nin \cdot \Nout$.

Now, let us use three differentiable masks $m_\mathrm{res}$, $m_\Nin$ and $m_\Nout$ to mask the resolution, input and output channels of the linear layer. The total number of computations is now given by:
\begin{equation*}
\setlength{\abovedisplayskip}{4pt}
\setlength{\belowdisplayskip}{4pt}
   \gC_{\mathrm{lin, masked}} =  \mathrm{size}(m_\mathrm{res}) \cdot \mathrm{size}(m_\Nin) \cdot \mathrm{size}(m_\Nout).\nonumber
\end{equation*}
Since the size of the masks is now involved in the computation of the operations required, we can utilize it as an additional source of feedback to update the masks by making the function $\mathrm{size}$ differentiable with regard to the mask parameters. The same concept is used to calculate the cost of other layers based on the size of the masks. A summary of these costs can be found in Appx.~\ref{appx:masking_other_layers}.

\textbf{Effect of the depth mask.} To take into account the effect of the depth mask, we use it to determine the number of residual blocks in the network. If the number of operations of a network with $\mathrm{D}$ residual blocks is denoted as $\gC_\mathrm{net, D}$, the complexity of a network with masked depth is given by $\gC_\mathrm{net, \mathrm{size}(m_\mathrm{depth})}$ with $\mathrm{size}(m_\mathrm{depth})$ the size of the depth mask.

\textbf{Computing the $\mathrm{size}$ of the masks.} The size of a mask can be calculated in a differentiable manner by determining the length of the mask in continuous space and using that length to estimate the change in size of the corresponding network dimension. Specifically, the length at a time $t$ is $2 x_{T_m}^{t}$ and $x_{T_m}^{t} {-} x_\mathrm{min}$, for Gaussian and Sigmoid masks, respectively (see Fig.~\ref{fig:diff_masks}). For some initial $x^{0}_{T_m}$ and corresponding initial length $\Nt$, the size of a Gaussian and a Sigmoid mask at time $t$ is respectively:
\begin{center}
\vspace{-5mm}
    \begin{tabular}{p{5cm}p{6cm}}
    \begin{equation}
        \mathrm{size}(m_\mathrm{gauss}) = \tfrac{2 x^{t}_{T_m}}{2 x^{0}_{T_m}} \Nt, \label{eq:size_gauss}
    \end{equation}
    &
    \begin{equation}
         \mathrm{size}(m_\mathrm{sigm}) = \tfrac{x^{t}_{T_m} - x_\mathrm{min}}{ x^{0}_{T_m} - x_\mathrm{min}} \Nt.\label{eq:size_sigm}
    \end{equation}
    \end{tabular}
    \vspace{-5mm}
\end{center}
\textbf{Computational constraints as an additional loss.} Let $\gC_{\mathrm{curr}}$ be the current complexity of the network and $\gC_{\mathrm{target}}$ be the desired target complexity. We define the computational loss $\gL_\mathrm{comp}$ as the relative $\mathcal{l}^2$ difference between the relative complexity of the current network and the target:
\begin{equation}
\setlength{\abovedisplayskip}{4pt}
\setlength{\belowdisplayskip}{4pt}
    \mathcal{l}^{2}\left(\tfrac{\gC_{\mathrm{curr}}}{\gC_{\mathrm{target}}}, 1\right) = \left\| \tfrac{\gC_{\mathrm{curr}}}{\gC_{\mathrm{target}}} - 1.0 \right\|^2_2.
\end{equation}
This form has two advantages over the alternative form $\mathcal{l}^2(\gC_{\mathrm{curr}}, \gC_{\mathrm{target}})$. It (\textit{i}) prevents overflow that might occur when comparing large values --$\gC_{\mathrm{curr}}$ and $\gC_{\mathrm{target}}$ may easily be of order $1\mathrm{e}^{10}$--, and (\textit{ii}) allows for consistent tuning of $\lambda$ for different tasks and complexities. In the alternative form $\mathcal{l}^2(\gC_{\mathrm{curr}}, \gC_{\mathrm{target}})$, $\lambda$ might need to be tuned independently for different complexity regimes.

\vspace{-3mm}
\section{Experiments}
\vspace{-2mm}
\begin{table*}
\centering
    \begin{minipage}{\textwidth}
    \centering
    \caption{Performance on the LRA benchmark. $\times$ denotes random guessing. Highest per-section scores are in bold and the overall best scores are underlined. For DNArch, values in parenthesis indicate the computational cost of the architecture relative to the target complexity.}
    \label{tab:lra_results}
    \vspace{-2.5mm}
    \begin{small}
    \scalebox{0.8}{
    \begin{tabular}{llllllll}
    \toprule
    \sc{Model} & \sc{ListOps}  & \sc{Text} &  \sc{Retrieval} & \sc{Image} & \sc{Pathfinder} & \sc{Path-X} & \sc{Avg.} \\
     \midrule
    Transformer \cite{vaswani2017attention} & \textbf{36.37} & 64.27 &  57.46 & 42.44 & 71.40 & $\times$& 53.66\\
    Reformer \cite{kitaev2020reformer} & 37.27 & 56.10 & 53.40 & 38.07 & 68.50 & $\times$& 50.56\\
    Performer \cite{choromanski2020rethinking} & 18.01 &\textbf{ 65.40} & 53.82 & \textbf{42.77} & \textbf{77.05}  & $\times$& 51.18 \\
    BigBird \cite{zaheer2020big} & 36.05 & 64.02 & \textbf{59.29} & 40.83 & 74.87  & $\times$& \textbf{54.17} \\
    \midrule
    Mega ($\mathcal{O}(L^2)$) \cite{ma2022mega} & \textbf{\underline{63.14}} & \underline{\textbf{90.43}} & \textbf{91.25} & \textbf{\underline{90.44}} &  \textbf{\underline{96.01}} & \textbf{97.98} & \textbf{\underline{88.21}}\\
    Mega-chunk ($\mathcal{O}(L)$) \cite{ma2022mega} & 58.76 & 90.19 & 90.97 &  85.80 & 94.41 & 93.81 &  85.66 \\
    \midrule
    S4D \cite{gu2022parameterization} & 60.47 & 86.18 & 89.46 & 88.19 & 93.06 & 91.95 & 84.89\\
    S4 \cite{gu2021efficiently} & 59.60 & 86.82 & 90.90 & \textbf{88.65} & 94.20 & 96.35 & 86.09\\
    S5 \cite{smith2022simplified} & \textbf{61.50} & \textbf{89.31} & \textbf{\underline{91.40}} & 88.00 & \textbf{95.33} & \textbf{\underline{98.58}} & \textbf{87.35}\\
    \midrule
        FNet \cite{lee2021fnet} &  35.33 & 65.11 & 59.61 & 38.67 & 77.80 & $\times$ & 54.42\\
    Luna-256 \cite{ma2021luna} & 37.25 & 64.57 & 79.29 & 47.38 & 77.72 & $\times$ & 59.37 \\
    CCNN$_{4, 140}$ \cite{romero2022towards} & \textbf{44.85} & \textbf{83.59}  & $\times$ & \textbf{87.62} & \textbf{91.36} & $\times$ & \textbf{76.86} \\
    \midrule
    \midrule
    CCNN$_{4, 140}$ (Global Kernels) & 55.65 & 87.80 & 90.55 & 85.51 & 94.26 & 91.15 & 84.15\\
    $\mathrm{DNArch}_\mathrm{K}$(CCNN$_{4, 140}$) & 59.90 & 88.28& 90.66  & 86.07 & 93.46 & 89.93 & 84.72\\
    $\mathrm{DNArch}_\mathrm{K,R}$(CCNN$_{4, 140}$) & 60.15$_{(0.80\times)}$ & 88.50$_{(0.75\times)}$ & 91.08$_{(0.78\times)}$  & 86.55$_{(0.82\times)}$ & 94.05$_{(0.89\times)}$ & 91.15$_{(0.82\times)}$ & 85.25\\
    $\mathrm{DNArch}_\mathrm{K,R,W,D}$(CCNN$_{4, 140}$) & \textbf{60.55}$_{(1.01\times)}$ & \textbf{89.03}$_{(1.00\times)}$ & \textbf{91.22}$_{(1.02\times)}$ & \textbf{87.20}$_{(1.02\times)}$ & \textbf{94.95}$_{(1.00\times)}$ & \textbf{91.71}$_{(1.01\times)}$ & \textbf{85.78} \\
    \bottomrule
    \end{tabular}}
    \end{small}
    \end{minipage}%
    \vspace{-6mm}
\end{table*}
We evaluate DNArch on sequential and image datasets for classification and dense prediction tasks. On 1D, we use the Long Range Arena (LRA) benchmark \citep{tay2020long}, which includes six sequence modelling tasks with sequence lengths ranging from 1024 to over 16000. On 2D, we perform image classification on the CIFAR10 and CIFAR100 datasets \citep{krizhevsky2009learning} and report results on two dense prediction tasks from the NAS-Bench-360 benchmark \citep{tu2022bench}: {\tt DarcyFlow} \citep{li2020fourier} and {\tt Cosmic} \citep{zhang2020deepcr}. A detailed description of the datasets used can be found in Appx.~\ref{appx:datset_description}.

\textbf{Experimental setup.} We use two CCNNs of different capacity as base networks: a CCNN$_{4,140}$ --4 blocks, 140 channels, 200{\sc{k}} parameters--, and a CCNN$_{6,380}$ --6 blocks, 380 channels, 2{\sc{m}} parameters--, and use DNArch to learn their architectures. To understand the impact of learning each network component, we also report results learning some and none of the neural architecture components. 

\textbf{Mask configurations.} We initialize all the masks to match the architecture of the baseline CCNNs at the beginning of training. We use a Gaussian mask to learn kernel sizes as in FlexConv \citep{romero2021flexconv}, and Sigmoid masks to learn width, depth and downsampling. All masks use a threshold of $T_m{=}0.1$. All kernel masks are centered, i.e., $\mu{=}0$, and initialized to either be small or global, i.e., $\sigma \in [0.0325, 0.5]$. Resolution masks are initialized to weight the highest input frequency by $0.85$, and the width and depth masks are initialized to match the size of the base network's architecture. More information on hyperparameters, training regimes, and experimental settings can be found in Appx.~\ref{appx:exp_details}. 

\textbf{Notations.} We use $\mathrm{DNArch}$ as an operator acting on a base network and specify the learned components with indices $\mathrm{K, R, W, D}$ representing kernel sizes, downsampling, width and depth. $\mathrm{DNArch}_\mathrm{K}$(CCNN$_{4, 140}$) indicates using DNArch to learn only the kernel sizes of a CCNN$_{4, 140}$.

\subsection{Using DNArch without computational constraints}\label{sec:no_constr}
\vspace{-2mm}
First, we use DNArch to improve the expressiveness and computational efficiency of a CCNN$_{4, 140}$. We start using DNArch to learn the receptive field of all convolutional layers, and then we learn both the kernel sizes and downsampling layers to simultaneously improve the expressiveness and the computational efficiency of the CCNN$_{4,140}$. It is worth noting that in this scenario learning is solely driven by the objective loss $\gL_\mathrm{obj}$, i.e., the regularization term $\gL_\mathrm{comp}$ is not used. In addition, as we use Fourier convolutions, the learned kernel sizes do not impact computational efficiency.

\textbf{Results.} Except for {\tt PathFinder} and {\tt Path-X}, we find that using DNArch to learn kernel sizes consistently improves the accuracy of the base architecture (DNArch$_\mathrm{K}$ models in Tabs.~\ref{tab:lra_results}-\ref{tab:image_classif}). Interestingly, found DNArch architectures perform on par, and even surpass, architectures specifically designed for each tasks, e.g., S4 \citep{gu2021efficiently} for sequential tasks and NFOs \cite{li2020fourier} for PDEs on 2D with a remarkably lower number of trainable parameters. In contrast to DARTS methods, e.g., DASH \cite{shen2022efficient}, DNArch can be applied across \textit{all tasks} without the need to \textit{manually change} the base architecture. When additionally learning downsampling, we observe that DNArch finds high-performant architectures with improved computational efficiency (DNArch$_\mathrm{K, R}$ models in Tabs.~\ref{tab:lra_results}-\ref{tab:image_classif}). Interestingly, we observe that the found models often exhibit slight accuracy improvements. This is explained by low resolution kernels being easier to model and construct than higher resolution ones.
\begin{table}
\begin{minipage}{.48\textwidth}
  \centering
  \caption{Results on dense prediction tasks.}
    \label{tab:nas_bench_360}
    \begin{small}
    \scalebox{0.8}{
    \begin{tabular}{lll}
    \toprule
     \multirow{2}{*}{\sc{Model}} & \sc{DarcyFlow}  & \sc{Cosmic} \\
     &  rel. $\mathcal{l}^2$ loss & 1 - \sc{auroc} \\
     \midrule
     Expert* & \textbf{0.008} & \textbf{ 0.13}\\
     \midrule
     WRN \cite{zagoruyko2016wide} & 0.073 & 0.24 \\
     DenseNAS \cite{fang2020densely} & 0.100 & 0.38\\
     DARTS \cite{liu2018darts} & \textbf{0.026} & 0.229 \\
     Auto-DL \cite{liu2019auto} & 0.049 & 0.495 \\
     DASH \cite{shen2022efficient} & 0.060 & \textbf{0.190}\\
    \midrule
    \midrule
    CCNN$_{4, 140}$ (Global Kernels) &  0.002989 & 0.059\\
    $\mathrm{DNArch}_\mathrm{K}$(CCNN$_{4, 140}$) & 0.002970 & 0.058\\
    $\mathrm{DNArch}_\mathrm{K,R}$(CCNN$_{4, 140}$) & 0.002929$_{(0.79\times)}$ & 0.056$_{(0.82\times)}$ \\
    $\mathrm{DNArch}_\mathrm{K,R,W,D}$(CCNN$_{4, 140}$) & \textbf{0.002285}$_{(1.01\times)}$ & \textbf{0.055}$_{(1.01\times)}$\\
    \midrule
    CCNN$_{6, 380}$ (Global Kernels) &  0.004521 & 0.059\\
    $\mathrm{DNArch}_\mathrm{K,R,W,D}$(CCNN$_{6, 380}$) & \textbf{\underline{0.001763}}$_{(1.00\times)}$ & \textbf{\underline{0.048}}$_{(1.00\times)}$ \\
    \bottomrule
    \multicolumn{3}{l}{$^*$ FNO \cite{li2020fourier} and deepCR \cite{zhang2020deepcr}.}
    \end{tabular}}
    \end{small}
\end{minipage}\hfill
\begin{minipage}{0.48\textwidth}
    \centering
    \caption{Results on image classification tasks.}
    \label{tab:image_classif}
    \begin{small}
    \scalebox{0.8}{
    \begin{tabular}{lll}
    \toprule
     \sc{Model} & \sc{CIFAR10} & \sc{CIFAR100} \\
     \midrule
     WRN \cite{zagoruyko2016wide} &  - & \textbf{\underline{76.65}} \\
     DenseNAS \cite{fang2020densely} & - & 74.51 \\
     DARTS \cite{liu2018darts} & - & 75.98 \\
     Auto-DL \cite{liu2019auto} & - & - \\
     DASH \cite{shen2022efficient} & - & 75.63 \\
    \midrule
    \midrule
    CCNN$_{4, 140}$ (Global Kernels) & 90.52 & 64.72\\
    $\mathrm{DNArch}_\mathrm{K}$(CCNN$_{4, 140}$) & 92.51 & 69.01\\
    $\mathrm{DNArch}_\mathrm{K,R}$(CCNN$_{4, 140}$) & 92.77$_{(0.82\times)}$ & 68.96$_{(0.85\times)}$  \\
    $\mathrm{DNArch}_\mathrm{K,R,W,D}$(CCNN$_{4, 140}$) & \textbf{93.47}$_{(1.01\times)}$ & \textbf{72.98}$_{(1.03\times)}$  \\
    \midrule
    CCNN$_{6, 380}$ (Global Kernels) & 94.18 & 72.29\\
    $\mathrm{DNArch}_\mathrm{K,R,W,D}$(CCNN$_{6, 380}$) & \textbf{\underline{95.03}}$_{(1.00\times)}$ & \textbf{76.37}$_{(1.02\times)}$  \\
    \bottomrule
    \end{tabular}}
    \end{small}
\end{minipage}
\vspace{-6mm}
\end{table}
\vspace{-3mm}
\subsection{Using DNArch under computational constraints}
\vspace{-2mm}
Next, we utilize DNArch to learn entire convolutional architectures that respect a predefined computational budget. To this end, we start with base  CCNN$_{4,140}$ and CCNN$_{6,380}$ networks, and allow DNArch to learn their width, depth, kernel sizes and downsampling. We define the target complexity $\mathcal{L}_{\mathrm{comp}}$ as the complexity of the base CCNN networks. In other words, we use DNArch to find better convolutional architectures of computational complexity roughly equal to that of the base networks.

\textbf{Results.} Our results (DNArch$_\mathrm{K, R, W, D}$ models in Tabs.~\ref{tab:lra_results}-\ref{tab:image_classif}) show that DNArch finds neural architectures that achieve higher accuracy than the base CCNN networks while keeping the same computational complexity. In addition, we observe that learning more neural architecture components consistently leads to better results, therefore supporting the claim that using gradient-steered architectures can be more beneficial than using handcrafted ones. Furthermore, we observe that using base architectures with larger complexity and capacity consistently leads to better results. This result is encouraging for the application of DNArch to large architectures, e.g., LLMs \cite{brown2020language, chowdhery2022palm}.

\begin{wrapfigure}{r}{0.42 \textwidth}
    \centering
    \vspace{-5mm}
    \includegraphics[width=0.42\textwidth]{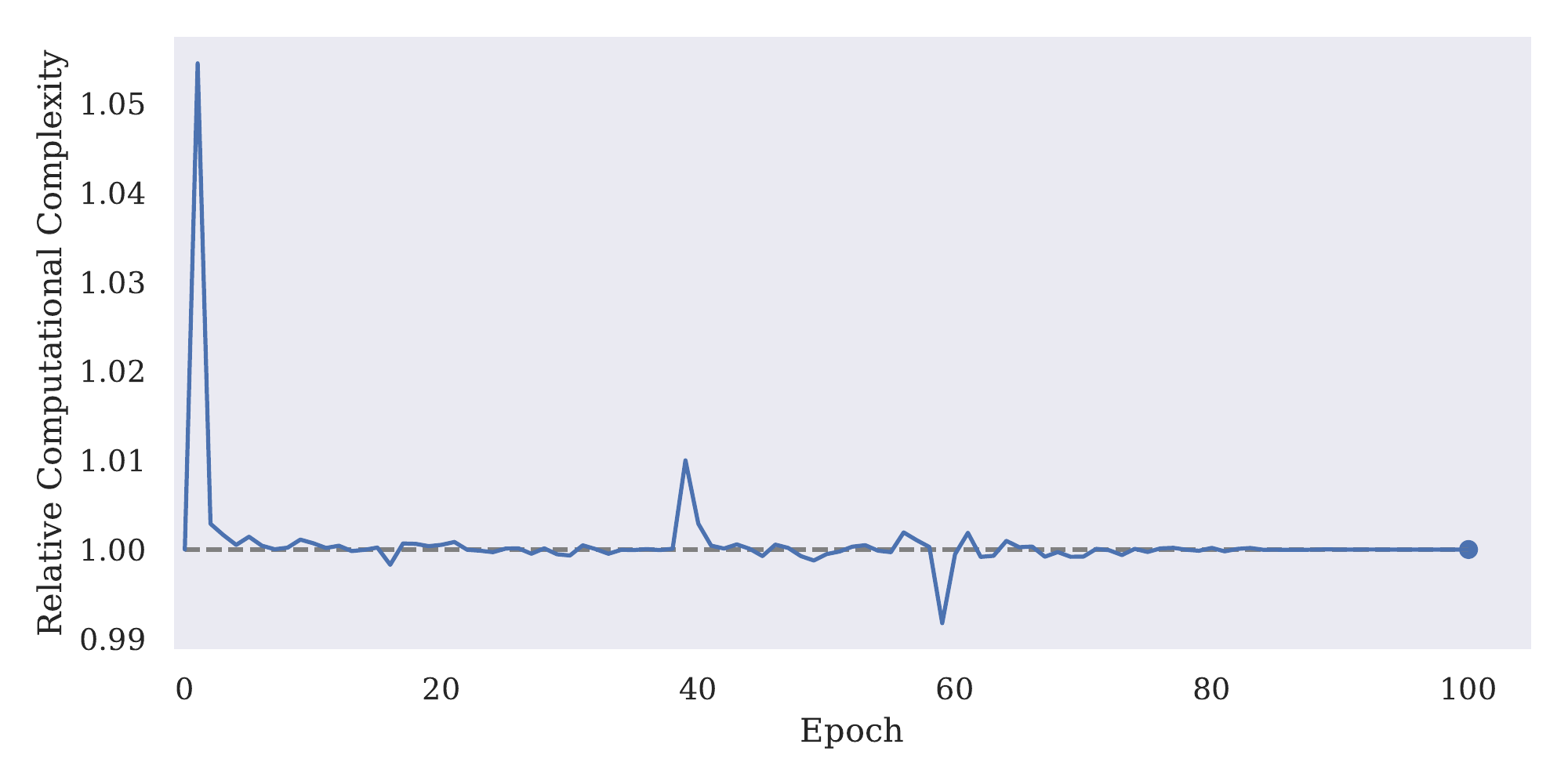}
    \vspace{-6mm}
    \caption{Relative complexity during the course of training on the {\tt Text} task. This behavior is consistent across all tasks.
    \vspace{-5mm}}
    \label{fig:rel_complexity}
\end{wrapfigure}

\newpage
\textbf{Computational complexity of DNArch.} To assess the applicability of DNArch, it is important to analyze its computational overhead. To this end, we analyze the behavior of the relative complexity ($\gC_\mathrm{curr}/\gC_\mathrm{target}$) during training (Fig.~\ref{fig:rel_complexity}). Interestingly, we observe that the theoretical complexity of candidate architectures $\gC_\mathrm{curr}$ stays close to the target complexity $\gC_\mathrm{target}$ \textit{during the whole training}. This indicates that: (\textit{i}) DNArch only searches among architectures that share the target computational complexity, and that (\textit{ii}) the computational overhead of DNArch is \textit{negligible}. As a result, the cost of using DNArch on top of a CCNN is comparable to the cost of training the base CCNN network. Note that, for the experiments in Sec.~\ref{sec:no_constr}, the cost of training can be even lower than that of the base network since the complexity of found architectures are up to $25\%$ faster.
\vspace{-3mm}
\subsection{Architectures found by DNArch} 
\vspace{-2mm}
The architectures found by DNArch are listed in Tabs.~\ref{tab:found architectures_lra}-\ref{tab:found architectures_2d}. Interestingly, we observe that found architectures are very diverse, even within each architecture. For instance, some residual blocks have a bottleneck structure, some an expanded structure, and others have monotonically decreasing or increasing widths. Interestingly, the resolution of found architectures for classification tasks, e.g., {\tt Text}, often follow the style of U-Nets, and not the monotonically decreasing pattern commonly seen in handcrafted networks. On dense tasks, we observe architectures that resemble U-Nets, and even concatenated U-Nets, e.g., the 1.5$\times$ U-Net like architecture found for the {\tt Cosmic} task. 

The kernel sizes found by DNArch are also very diverse. In 1D tasks, found kernels are often large, which would make them parameter intensive with traditional parameterizations. In 2D tasks, we often see rectangular kernels that do not follow a monotonic pattern of increasing or decreasing sizes. Instead, found architectures often perform interleaved low-level and high-level feature extraction.
\vspace{-3mm}
\section{Limitations}\label{sec:limitations}
\vspace{-3mm}
\textbf{Training on {\tt TPU} requires static shapes.} We train our models on {\tt TPUs}, a type of accelerator that requires a static computational graph derived for specific input and network shapes via the {\tt XLA} (Accelerated Linear Algebra) compiler. As a result, {\tt TPUs} do not support operations that change the shapes of arrays during training. This means that on {\tt TPUs}, DNArch can only perform masking modifications to the network during training, i.e., setting certain channels to zero but still computing their outputs. At inference, however, the masks are fixed. Consequently, we can effectively trim unused values to remove useless computations in a way that is compatible with {\tt XLA}. It is important to note that this limitation is solely an implementation issue caused by nature of {\tt TPUs}' hardware and can be avoided by using libraries and hardware that support dynamic computational graphs, e.g., {\tt PyTorch} and {\tt GPUs}. While our results were obtained using {\tt TPUs}, we also provide a {\tt PyTorch} implementation that avoids this issue, making it more flexible and accessible, especially in scenarios where one needs to keep candidate networks close to the target complexity $\gC_\mathrm{target}$ during training.

\textbf{DNArch requires instantiating the largest possible architecture.} While masking weights through a gradient update is straightforward, increasing the number of active weights requires those weights to be instantiated in memory. This means that even with dynamic computational graphs, it is necessary to instantiate the largest possible architecture learnable by DNArch in memory. To overcome this limitation, we set the maximum kernel size to the size of the input, and limit the maximum network size along the depth and width dimensions to double the number of blocks and channels of the base network. While this trick allows DNArch to easily shrink and grow representations within that range, this restricts the potential sizes of optimal architectures and can restrict the applicability of DNArch to very large models, e.g., LLMs \cite{brown2020language, chowdhery2022palm}, which can have billions of weights.
\vspace{-4mm}
\section{Outlook and future work}
\vspace{-3mm}
\textbf{Input-dependent neural architectures.}
In this work, the mask parameters are constant for all inputs within a task. An alternative approach could use an additional neural network \mlp$_\mathrm{mask}$ to predict the mask parameters based on context, e.g., the current input, current task, etc. This would enable the creation of context-dependent neural architectures such as early-exit systems \cite{teerapittayanon2016branchynet, ghodrati2021frameexit, schuster2022confident}, but where the whole network architecture is context-dependent. Consequently, resulting architectures would providing finer control of per-sample / per-modality complexity than existing methods. 

\textbf{Dynamic weighting of $\gL_\mathrm{comp}$ during training.}
DNArch explores architectures with complexity similar to target complexity throughout training. This results from using a constant $\lambda$ in Eq.~\ref{eq:optim_loss}. Alternatively, one could use a dynamic value of $\lambda$ during training to induce a different training behavior. For example, gradually increasing $\lambda$ would allow DNArch to explore architectures with larger complexity at first, and progressively encourage it to converge to networks with the desired target complexity. Such a weighting scheduling of $\lambda$ could lead DNArch to find better architectures.

\textbf{Training DNArch with additional / multiple constraints.} Here, we only consider computational complexity as a constraint when training with DNArch. However, other properties such as memory efficiency, hardware-awareness and robustness are equally important. Designing regularization terms that encourage other properties in DNArch as well as exploring how different properties can be optimized in unison are important directions for further research.




\bibliographystyle{abbrvnat}
\bibliography{references}

\newpage
\appendix
\section*{\Large Appendix}

\section{Architectures Found by DNArch}
\begin{table}[h]
    \centering
    \caption{Architectures found by DNArch on LRA with the target complexity of a CCNN$_4,140$.}
    \label{tab:found architectures_lra}
    \begin{small}
    \scalebox{0.8}{
    \begin{tabular}{ccccc}
    \toprule
     \multirow{2}{*}{\sc{Task}} &\multirow{2}{*}{\sc{Depth}} & \multirow{2}{*}{\sc{Kernel Size}} & \multirow{2}{*}{\sc{Resolution}} & \sc{Width} \\
     & & & & [$\Nin$, $\Nt_\mathrm{mid}$, $\Nout$] \\
     \midrule
     \multirow{8}{*}{\rotatebox{90}{\sc{ListOps}}} & \multirow{8}{*}{8} & 266 & 2048 & [150 189 145]\\
     & & 569 & 632 & [150 168 168] \\
     & & 1401 & 1416 & [176 186 162]\\
     & & 310 & 310 & [166 175 153]\\
     & & 213 & 213 & [154 159 163]\\
     & & 12  & 301 & [168 128 162]\\
     & & 5 & 250 & [170 158 153]\\
     & & 24 & 502 & [153 171 165]\\
     \midrule
    \multirow{8}{*}{ \rotatebox{90}{\sc{Text}}} & \multirow{8}{*}{8} &  445 & 2284 & [180 217 205]\\
     & & 691 & 2939 & [208 176 153]\\
     & & 1420 & 1420 & [152 152 120] \\
     & & 415 & 1313 & [120 120 147]\\
     & & 1467 & 1467 & [147 118 135]\\
     & & 52 & 594 & [134 173 153]\\
     & & 101 & 932 & [150 156 183]\\
     & & 149 & 1036 & [180 92 192] \\
    \midrule
    \multirow{8}{*}{\rotatebox{90}{\sc{Retrieval}}} & \multirow{8}{*}{8} & 2 & 1913 & [29 33 172] \\
    & & 136 & 2058 & [184 174 183] \\
    & & 1013 & 2363 & [205 171 161] \\
    & & 1446 & 2724 & [188 164 115] \\
    & & 7 & 2604 & [29 29 163]\\
    & & 1 & 2756 & [29 35 154] \\
    & & 6 & 3545 & [71 110 147]\\
    & & 1 & 3899 & [71 88 137] \\
    \midrule
    \multirow{8}{*}{\rotatebox{90}{\sc{Image}}} & \multirow{8}{*}{8} & 203 & 1024 & [118 155 147]\\
    & & 279 & 1024 & [146 172 164]\\
    & & 219 & 486 & [173 166 196]\\
    & & 308 & 308 & [199 197 196]\\
    & & 144 & 144 & [207 197 92]\\
    & & 8 & 125 & [106 29 75]\\
    & & 30 & 96 & [78 28 110]\\
    & & 40 & 126 & [104 51 104]\\
    \midrule
    \multirow{8}{*}{\rotatebox{90}{\sc{PathFinder}}} & \multirow{8}{*}{8} & 195 & 1024 & [109 140 171]\\
    & & 493 & 770 & [171 168 158] \\
    & & 418 & 507 & [144 183 170]\\
    & & 318 & 318 & [173 187 178]\\
    & & 236 & 236 & [182 162 160]\\
    & & 231 & 231 & [161 121 103]\\
    & & 8 & 251 & [105 47 210]\\
    & & 4 & 253 & [116 29 188]\\
    \midrule
    \multirow{5}{*}{\rotatebox{90}{\sc{Path-X}}} & \multirow{5}{*}{5}  & 2484 & 15331 & [280 174 157] \\
    & & 7204 & 7204 & [177 280 159]\\
    & & 3669 & 3772 & [167 280 98]\\
    & & 2323 & 5496 & [123 164 164]\\
    & & 513 & 4768 & [136 128 195]\\
    \bottomrule
    \end{tabular}}
    \end{small}
\end{table}

\begin{table}
    \centering
    \caption{Architectures found by DNArch on 2D datasets with the target complexity of a CCNN$_{4,140}$.}
    \label{tab:found architectures_2d_ccnn4140}
    \begin{small}
    \scalebox{0.8}{
    \begin{tabular}{ccccc}
    \toprule
     \multirow{2}{*}{\sc{Task}} &\multirow{2}{*}{\sc{Depth}} & \sc{Kernel Size} & \sc{Resolution} & \sc{Width} \\
     & & [\sc{y} \sc{x}] & [\sc{y} \sc{x}] & [$\Nin$, $\Nt_\mathrm{mid}$, $\Nout$] \\
     \midrule
    \multicolumn{5}{c}{\sc{Image Classification Tasks}} \\
    \midrule
    \multirow{8}{*}{\rotatebox{90}{\sc{Cifar10}}} & \multirow{8}{*}{8} & [9 7]  & [32 32] & [142 139 145]\\
    & & [12 8] & [32 32] & [145 160 157]\\
    & & [25 7] & [32 20] & [158 186 182] \\
    & & [9 10] & [9 15] & [186 208 168]\\
    & & [1 13] & [5 15] & [169 177 150] \\
    & & [1 10] & [6 11] & [151 139 156]\\
    & & [5 1] & [15 4] & [154 115 110] \\
    & & [6 5] & [11 7] & [108 41 166]\\
    \midrule
    \multirow{8}{*}{\rotatebox{90}{\sc{Cifar100}}} & \multirow{8}{*}{8} & [13 7] & [32 32] & [104 107 116] \\
    & & [6 10] & [32 32] & [114 134 134]\\
    & & [11 8] & [22 22] & [139 192 166] \\
    & & [13 7] & [16 18] & [173 201 197]\\
    & & [8 12] & [10 12] & [205 251 51]\\
    & & [1 1] & [8 9] & [62 56 157] \\
    & & [5 9] & [8 10] & [162 175 254] \\
    & & [8 7] & [9 9] & [280 280 280] \\
    \midrule
        \multicolumn{5}{c}{\sc{Dense Tasks}} \\
    \midrule
    \multirow{3}{*}{\rotatebox{90}{\makecell{\sc{Darcy}\\ \sc{Flow}}}}& \multirow{3}{*}{3} & [43 38] & [80 72] & [156 280 107] \\
    & & [22 22] & [22 22] & [180 204 78] \\
    & & [76 76] & [85 85] & [280 280 50] \\
    \midrule
    \multirow{6}{*}{\rotatebox{90}{\sc{Cosmic}}} & \multirow{6}{*}{6} & [94 111] & [128 128] & [18 110 23] \\
    & & [2 13] & [20 45] & [186 207 139] \\
    & & [129 129] & [129 129] & [126 265 100] \\
    & & [129 121] & [129 129] & [78 105 59] \\
    & & [90 89] & [129 129] & [57 201 197] \\
    & & [76 74] & [76 74] & 202 145 216] \\
    \bottomrule
    \end{tabular}}
    \end{small}
\end{table}

\begin{table}
    \centering
    \caption{Architectures found by DNArch on 2D datasets with the target complexity of a CCNN$_{6,380}$.}
    \label{tab:found architectures_2d}
    \begin{small}
    \scalebox{0.8}{
    \begin{tabular}{ccccc}
    \toprule
     \multirow{2}{*}{\sc{Task}} &\multirow{2}{*}{\sc{Depth}} & \sc{Kernel Size} & \sc{Resolution} & \sc{Width} \\
     & & [\sc{y} \sc{x}] & [\sc{y} \sc{x}] & [$\Nin$, $\Nt_\mathrm{mid}$, $\Nout$] \\
     \midrule
    \multicolumn{5}{c}{\sc{Image Classification Tasks}} \\
    \midrule
    \multirow{12}{*}{\rotatebox{90}{\sc{Cifar10}}} & \multirow{12}{*}{12} & [4 7]  & [32 32] & [380 328 384]\\
    & & [9 10] & [32 32] & [384 371 393]\\
    & & [12 6] & [32 32] & [392 361 391] \\
    & & [20 6] & [32 32] & [388 370 421]\\
    & & [10 11] & [23 26] & [421 417 486] \\
    & & [11 11] & [12 22]  & [496 444 479]\\
    & & [1 11] & [6 11] & [493 482 304] \\
    & & [1 6] & [5 21] & [211 78 384]\\
    & & [29 4] & [32 4] & [363 459 280] \\
    & & [18 15] & [18 15] & [277 394 67]\\
    & & [1 1] & [4 4] & [111 109 361] \\
    & & [4 3] & [21 15] & [121 374 449]\\
    \midrule
    \multirow{12}{*}{\rotatebox{90}{\sc{Cifar100}}} & \multirow{12}{*}{12} & [8 9]  & [32 32] & [343 275 354]\\
    & & [12 10] & [32 32] & [351 316 397]\\
    & & [11 10] & [32 32] & [495 355 420] \\
    & & [18 12] & [29 21] & [421 498 419]\\
    & & [11 15] & [27 24] & [432 449 407] \\
    & & [19 8] & [25 20]  & [412 419 413]\\
    & & [11 10] & [12 23] & [423 454 600] \\
    & & [8 8] & [8 9] & [709 685 416]\\
    & & [5 7] & [5 8] & [419 311 446] \\
    & & [8 4] & [8 4] & [446 433 389]\\
    & & [6 4] & [6 4] & [386 501 570] \\
    & & [8 9] & [8 9] & [568 453 655]\\
    \midrule
        \multicolumn{5}{c}{\sc{Dense Tasks}} \\
    \midrule
    \multirow{7}{*}{\rotatebox{90}{\makecell{\sc{Darcy}\\ \sc{Flow}}}}& \multirow{7}{*}{7} & [54 49] & [54 49] & [435 428 289] \\
    & & [43 47] & [70 72] & [499 393 284] \\
    & & [65 69] & [85 85] & [496 434 281] \\
    & & [67 66] & [85 85] & [323 412 275] \\
    & & [85 85] & [85 85] & [319 369 271] \\
    & & [85 85] & [85 85] & [306 379 258] \\
    & & [68 68] & [85 85] & [521 435 271] \\
    \midrule
    \multirow{12}{*}{\rotatebox{90}{\sc{Cosmic}}} & \multirow{12}{*}{12} & [35 32] & [35 33] & [146 236 272] \\
    & & [11 21] & [95 72] & [170 284 319] \\
    & & [44 24] & [128 128] & [141 339 388] \\
    & & [23 41] & [128 128] & [385 407 361] \\
    & & [28 27] & [128 128] & [351 279 356] \\
    & & [21 19] & [128 128] & [354 362 310] \\
    & & [29 24] & [128 128] & [310 351 466]] \\
    & & [18 25] & [128 128] & [396 292 183] \\
    & & [57 16] & [128 128] & [179 210 580] \\
    & & [50 11] & [127 77] & [273 250 63] \\
    & & [18 12] & [89 67] & [347 400 77] \\
    & & [22 23] & [97 79] & [171 241 79] \\
    \bottomrule
    \end{tabular}}
    \end{small}
\end{table}




\section{Learning downsampling in the spatial Domain}\label{appx:downsample_no_fourier}
\vspace{-1mm}
Differentiable Masking learns downsampling by multiplying the spectrum $\tilde{f}{=}\gF[f]$ of a signal $f$ with a differentiable mask $m(\cdot\ ; \boldsymbol{\theta})$, and cropping the output above the cutoff frequency of the mask $\omega_\mathrm{max}$ next. However, it is not strictly necessary to perform this operation in the Fourier domain. An equivalent downsampling can also be learned directly in the spatial domain.

The \textit{Fourier convolution theorem} states that the spatial convolution is equivalent to a pointwise multiplication in the Fourier domain. However, this equivalence works in both directions. That is, we can equivalently say that the pointwise multiplication on the Fourier domain is equal to a convolution on the spatial domain. Consequently, we can represent the pointwise multiplication of the spectrum of the input $\gF[f]$ and the differentiable mask $m(\cdot\ ; \boldsymbol{\theta})$ as the convolution of their inverse Fourier transforms. Formally:
\begin{align}
    \tilde{f} \cdot m(\cdot\ ; \boldsymbol{\theta}) &= \gF\left[ \gF^{-1} \left[\tilde{f} \right] * \gF^{-1}[m(\cdot\ ; \boldsymbol{\theta})] \right] \nonumber \\
   & = \gF\left[ \gF^{-1} \left[\gF[f] \right] * \gF^{-1}[m(\cdot\ ; \boldsymbol{\theta})] \right] \nonumber \\
    & =  \gF\left[f* \gF^{-1}[m(\cdot\ ; \boldsymbol{\theta})] \right] \label{eq:downsampling_spatial}
\end{align}
In other words, we can perform the same operation in the spatial domain by convolution the original input signal $f$ with the inverse Fourier transform of the mask $m(\cdot\ ; \boldsymbol{\theta})$.

\textbf{Defining the output resolution.} Eq.~\ref{eq:downsampling_spatial} defines how low-pass filtering can be performed on the spatial domain, but it does not provide information regarding the final resolution of the operation. To derive the resolution of the output, we can simply use Eqs.~\ref{eq:size_gauss},~\ref{eq:size_sigm} to analytically derive the size of the mask. Once the size of the mask is derived, we can simply take the downsampled signal --after using Eq.~\ref{eq:downsampling_spatial}--, and downsample it to match the size of the mask.
\vspace{-3mm}
\section{Computational complexity of masked network components}\label{appx:masking_other_layers}
\vspace{-1mm}
In this section, we derive the computational complexity of all layers used in the CCNN architecture with and without the use of masks. The calculation of these complexities follows the same reasoning as the pointwise linear layer provided as example in the main text.

With $\Lt$, $\Nin$ and $\Nout$ the length, number of input channels and number of output channels of a given layer, and $\mathrm{size}(m_\mathrm{res})$, $\mathrm{size}(m_\Nin)$, $\mathrm{size}(m_\Nout)$ the size of the masks along the corresponding dimensions, the complexity of the layers used in the CCNN architectures are given by:

\textbf{Pointwise linear layer:}
\begin{align*}
&\gC_{\mathrm{lin}}(f) = \Lt \cdot \Nin \cdot \Nout \\
&  \gC_{\mathrm{lin, masked}} =  \mathrm{size}(m_\mathrm{res}) \cdot \mathrm{size}(m_\Nin) \cdot \mathrm{size}(m_\Nout)
\end{align*}
\textbf{Fourier convolution:}
\begin{align*}
    &\gC_{\gF\mathrm{conv}} = \Lt  \log \left( \Lt \right) \\
    &\gC_{\gF\mathrm{conv,masked}} = \mathrm{size}(m_\mathrm{res}) \log \left(\mathrm{size}(m_\mathrm{res})\right)
\end{align*}
\textbf{Pointwise operations --GELU, DropOut, etc.--:}
\begin{align*}
    &\gC_\mathrm{pointwise} = \Lt \cdot \Nin \\
    &\gC_\mathrm{pointwise, masked} = \mathrm{size}(m_\mathrm{res})\cdot \mathrm{size}(m_\Nin)
\end{align*}

\vspace{-3mm}
\section{Dataset descriptions}\label{appx:datset_description}
\vspace{-1mm}
\subsection{The Long Range Arena benchmark}\label{appx:long_range_arena}
\vspace{-1mm}
The Long Range Arena \cite{tay2020long} consists of six sequence modelling tasks with sequence lenghts ranging from 1024 to over 16000. It encompasses modalities and objectives that require similarity, structural, and visuospatial reasoning. We follow the data preprocessing steps of \citet{gu2021efficiently}, which we also include here for completeness.

\textbf{ListOps.} An extended version of the dataset presented by \citet{nangia2018listops}. The task involves computing the integer result in the range zero to nine of a mathematical expression represented in prefix notation with brackets, e.g., [MAX29[MIN47]0] $\rightarrow$ 9. Characters are encoded as one-hot vectors, with 17 unique values possible (opening brackets and operators are grouped into a single token). The sequences are of unequal length. Hence, the end of shorter sequences is padded with a fixed indicator value to a maximum length of 2048. The task has 10 different classes
representing the possible integer results of the expression. It consists of 96{\sc{k}} training sequences, 2{\sc{k}}
validation sequences, and 2{\sc{k}} test sequences. No data normalization is applied.

\textbf{Text.} Based on the IMDB sentiment analysis dataset presented by \citet{maas2011learning}, the task is to classify movie reviews as having a positive or negative sentiment. The reviews are presented as a sequence of 129 unique integer tokens padded to a maximum length of 4096. The dataset contains 25{\sc{k}} training sequences and 25{\sc{k}} test sequences. No validation set is provided. No data normalization is applied.

\textbf{Retrieval.} Based on the ACL Anthology network corpus presented by \citet{radev2013acl}, the task is to classify whether two given textual citations are equivalent. To accomplish this, each citation is separately passed through an encoder, and passed to a final classifier layer. Denoting $X_1$ as the encoding for the first document and $X_2$ as the encoding for the second document, four features are created and concatenated together as:
\begin{equation*}
    X = [X_1, X_2, X_1 \times X_2, X_1 - X_2],
\end{equation*}
which are subsequently passed to a two layered MLP. The goal of the task is to evaluate how well the network can represent the text by evaluating if the two citations are equivalent or not. Characters are encoded into a one-hot vector with 97 unique values and sequences are padded to a maximum length of 4000. The dataset includes 147.086 training pairs, 18.090 validation pairs, and 17.437 test pairs. No normalization is applied.

\textbf{Image.} The Image task uses $32{\times}32$ images of the CIFAR10 dataset \cite{krizhevsky2009learning}. It views the images as sequences of length 1024 that correspond to a one-dimensional raster scan of the image. There are a total of 10 classes, 45{\sc{k}} training examples, 5{\sc{k}} validation examples and 10{\sc{k}} test examples. The RGB pixel values are converted to grayscale intensities and then normalized to have zero mean and unit variance across the entire dataset.

\textbf{PathFinder.} Based on the PathFinder challenge introduced by \citet{linsley2018learning}, the task presents a $32{\times}32$ grayscale image with an start and an end point depicted as small circles. The task is to classify whether there is a dashed line (or path) joining the start and end points while presenting the input as a one-dimension raster scan of the image, alike the {\tt Image} task. The dataset includes 160{\sc{k}} training examples, 20{\sc{k}}validation examples and 20{\sc{k}} test examples. The input data is normalized to be in the range [-1, 1].  

\textbf{Path-X.} Path-X is an \enquote{extreme} version of the {\tt PathFinder} dataset, in which the input images are of size $128{\times}128$. As a result, the input sequences are sixteen times longer with a total length of $16384$. Aside from this difference, the task is identical to the {\tt PathFinder} dataset.
\vspace{-3mm}
\subsection{Image classification datasets}
\vspace{-1mm}
\textbf{CIFAR10 and CIFAR100.} The CIFAR10 dataset \cite{krizhevsky2009learning}
consists of 60{\sc{k}} real-world 32${\times}$32 RGB images uniformly drawn from 10 classes divided into training
and test sets of 50{\sc{k}} and 10{\sc{k}} samples, respectively. The CIFAR100 dataset \cite{krizhevsky2009learning}
is similar to the CIFAR10 dataset, with the difference that the images are uniformly drawn from
100 different classes. For validation purposes, we divide the training dataset of both CIFAR10 and CIFAR100
into training and validation sets of 45{\sc{k}} and 5{\sc{k}} samples, respectively. Both datsets are normalized to have zero mean and unit variance across the entire dataset.
\vspace{-3mm}
\subsection{NAS-Bench-360}
\vspace{-1mm}
NAS-Bench-360 \cite{tu2022bench} is a benchmark suite to evaluate Neural Architecture Search methods beyond image classification. The benchmark is composed of ten tasks spanning a diverse array of application domains, datset sizes, problem dimensionalities, and learning objectives. In this work, we consider two tasks from the NAS-Bench-360 suite which require dense predictions: The {\tt DarcyFlow} \cite{li2020fourier} and {\tt Cosmic} \cite{zhang2020deepcr} datasets.

\textbf{DarcyFlow: Solving Partial Differential Equations.} DarcyFlow aims to solve Partial Differential Equation (PDE) by using neural networks as a replacement for traditional solvers. The input for this task is a $85{\times}85$ grid specifying the initial conditions and coordinates of a fluid, and the output is a 2D grid of the same dimensions representing the fluid state at a later time. The ground truth for this task is the result computed by a traditional solver, and the objective is to minimize the Mean Squared Error (MSE) between the predicted fluid state and the ground truth.

\textbf{Cosmic: Identifying Cosmic Ray Contamination.} The Cosmic task involves identifying and masking corruption caused by charged particles collectively referred to as \enquote{cosmic rays} on images taken from space-based facilities. It uses imaging data of local resolved galaxies collected from the Gubble Space Telescope. The input is an $128{\times}128$ image corresponding to the artifact of cosmic rays, and the output is a 2D grid of the same dimensions predicting whether each pixel in the input is an artifact of cosmic rays. We report the false-negative rate of identification results.
\vspace{-3mm}
\section{Experimental details}\label{appx:exp_details}
\vspace{-1mm}
\subsection{General remarks}
\vspace{-1mm}
\textbf{Code.} Our code is written in {\tt JAX} and our experiments are conducted on {\tt TPUs} and {\tt GPUs}. As outlined in the Limitations (Sec.~\ref{sec:limitations}), {\tt JAX} and {\tt TPU} training prevent DNArch from performing operations that change the dimensions of arrays during training. In addition to our {\tt JAX} implementation, we release a {\tt PyTorch} implementation of DNArch that supports these operations. This implementation makes DNArch more flexible and accessible, especially in scenarios where it is crucial to keep candidate networks close to the target complexity during the course of training.

\textbf{The Continuous CNN and the CCNN residual block.} The CCNN architecture is shown in Fig.~\ref{fig:ccnn_architecture}. It is composed by an {\tt Encoder}, a {\tt Decoder}, and a number of CCNN residual blocks {\tt ResBlock} \cite{knigge2023modelling}. The {\tt Encoder} is defined as a sequence of [{\tt PWLinear}, {\tt BatchNorm} \cite{ioffe2015batch}, {\tt GELU} \cite{hendrycks2016gaussian}] layers. For tasks dealing with text, we additionally utilize an Embedding layer mapping each token in the vocabulary to a vector representation of length equal to that used by \citet{gu2021efficiently} (see Appx.~\ref{appx:long_range_arena}).
For dense prediction tasks, the {\tt Decoder} is a {\tt{PWLinear}} layer, which is preceded by {\tt GlobalAvgPooling} for global prediction tasks.

\textbf{Batch Normalization in DNArch.} As the architecture is constantly changing during the search process, we use batch-specific statistics for batch normalization instead of the global moving average. This approach was adopted after early experiments showed that using the global moving average leads to a significant discrepancy in the behavior of the validation and training curves. Specifically, we observed that while the training curves were converging to a good solution, the validation curves resembled random predictions. This issue was resolved by deactivating the global moving average in Batch Normalization layers.

\textbf{Continuous convolutional kernels \mlp$_{\psi}$.} We parameterize our convolutional kernels as a 4-layer MLP with 128 hidden units and a Fourier Encoding \citep{tancik2020fourier} of the form $\gamma(\xv){=}[\cos(2\pi \omega_0 \mat{W} \xv]), \sin(2\pi \omega_0 \mat{W} \xv)]$, with $\Wm \in \sR^{\mathrm{D}\times 128}$ and $\omega_0$ a hyperparameter that acts as a prior on the frequency content of the kernels \cite{romero2021ckconv, sitzmann2020implicit}. In contrast to \citet{romero2021ckconv}, we utilize a single larger \mlp$_{\psi}$ to generate the kernels of the entire network. This allows the network \mlp$_{\psi}$ to administrate its capacity across all layers. Using different \mlp$_{\psi}$'s for each layer as \citet{romero2021ckconv} is inadequate in the learnable architectures setting as some layers can be entirely erased. With our proposed solution, the capacity of the otherwise zeroed-out \mlp$_{\psi}$ is used to generate the kernels of the remaining layers.

\textbf{Normalized relative positions.} Following \citet{romero2021ckconv, romero2021flexconv}, we normalize the coordinates going into \mlp$_{\psi}$ to lie in the space $[-1, 1]^{\Dt}$ for $\Dt$-dimensional kernels.

\textbf{Parameters and hyperparameters of the differentiable masks.} We learn some of the parameters of the masks, and leave the others constant or treat them as a hyperparameter. Specifically, for Gaussian masks, we only learn their width, i.e., $\sigma$, and fix its mean to zero. For Sigmoid masks, we learn their offset $\mu$ and treat their temperature $\tau$ as a hyperparameter. For more information regarding the values of $\tau$ used in our parameter tuning step, please refer to Appx.~\ref{appx:hyperparams}.

\textbf{Maximum and minimum allowable sizes for the learnable differentiable masks.} We define some minimum and maximum allowable sizes for the mask parameters, and reset them to these values after each training iteration if the updated parameter values lie outside that range. For the Gaussian mask, we constraint the minimum admissible value of $\sigma$ such that the length of the corresponding dimension never collapses to a value of $1$. This is to prevent the corresponding dimension to collapse such that it can grow afterwards if required. The minimum value depends on the resolution of the corresponding dimension, e.g., the maximum size of the convolutional kernel, and can be easily calculated with Eq.~\ref{eq:inv_gauss_mask}.

Note that the offset value of the Sigmoid mask $\mu$ could in principle assume any value in $\sR$. However, if not controlled, $\mu$ could become too small and mask all values along a particular dimension to zero. Similarly, if $\mu$ is too large, the gradient of the mask at all positions would become very small and it would difficult to update the mask. To avoid these situations, we define minimum and maximum values of $\mu$ such that for the lowest value, the mask at the lowest position is equal to $0.95$, and for the largest value, the mask at the highest position is equal $0.85$. These values are dependent on the value of the mask temperature $\tau$, and can be easily calculated with Eq.~\ref{eq:inv_sigm_mask}.

\textbf{Limiting the size of the mask to the maximum allowable ranges.} As outlined in the Limitations (Sec.~\ref{sec:limitations}), we must set a maximum allowable size for the width and depth of the network on {\tt JAX}. However, the maximum allowed value for the parameters of the masks (see previous paragraph) allows both masks to grow beyond the point on which the theoretical size of the masks is equal to the maximum allowable network size. For instance, for the maximum allowed parameter values of a Sigmoid mask, the last channel, i.e., the $280$-th channel, would be weighted by a factor of $0.85$. Consequently, the theoretical size of the mask as calculated by Eq.~\ref{eq:size_sigm} will be well beyond $280$. This value would lead to an unrealistic theoretical computational complexity that surpasses the real computational complexity the CCNN used.
\begin{table*}
    \centering
    \caption{Hyperparameters used for the experiments with the target complexity of a CCNN$_{4, 140}$.}
    \label{tab:hyperparams}
    \vspace{-2.5mm}
    \begin{small}
    \scalebox{0.8}{
    \begin{tabular}{lcccccccccc}
    \toprule
    \multirow{2}{*}{\sc{Dataset}} & \multirow{2}{*}{\sc{Epochs}} & \sc{Batch} & \sc{Learning} & \multirow{2}{*}{\sc{DropOut}} & \sc{Weight} & \multirow{2}{*}{$\omega_0$} & \multirow{2}{*}{$\lambda$} & \multirow{2}{*}{$\tau_\mathrm{resolution}$} & \multirow{2}{*}{$\tau_\mathrm{channel}$} & \multirow{2}{*}{$\tau_\mathrm{depth}$}\\
    & & \sc{Size} & \sc{Rate} & & \sc{Decay}\\
     \midrule
     \sc{ListOps} & 50 & 50 & 0.005 & 0.0 & 0.01 & 27.5{\sc{k}} & 5.0 & 50 & 25 & 8 \\
     \sc{Text}  & 100 & 50 & 0.02 & 0.2 & 0.01 & 19.5{\sc{k}} & 0.1 & 50 & 25 & 8 \\
     \sc{Retrieval} & 50 & 50 & 0.001 & 0.1 & 0.01 & 21.5{\sc{k}}  & 0.1 & 50 & 25 & 8 \\
     \sc{Image} & 210 & 50 & 0.02 & 0.1 & 0.001 & 12.5{\sc{k}} & 0.1 & 25 & 25 & 8 \\
     \sc{PathFinder} & 210 & 50 & 0.005 & 0.0 & 0.001 & 21.5{\sc{k}} & 0.1 & 50 & 25 & 8  \\
     \sc{Path-X} & 80 & 32 & 0.001 & 0.0 & 0.0 & 30{\sc{k}} & 0.1 & 100 & 25 & 8\\
     \midrule
     \sc{CIFAR10} & 210  & 50 & 0.01 & 0.1 & 0.01 & 21.5{\sc{k}} & 0.1 & 25 & 25 & 8 \\
     \sc{CIFAR100} & 210 & 50 & 0.01 & 0.0 & 0.01 & 6.5{\sc{k}} & 5.0 & 25 & 25 & 8\\
     \midrule
     \sc{DarcyFlow} & 310 & 8 & 0.02 & 0.0 & 0.0001 & 24.5{\sc{k}} & 0.1 & 50 & 25 & 8 \\
     \sc{Cosmic} & 310 & 8 & 0.02 & 0.3 & 0.0001 &5.5{\sc{k}} & 0.1 & 100 & 25 & 8 \\
    \bottomrule
    \end{tabular}}
    \end{small}
    \vspace{-4mm}
\end{table*}

\begin{table*}
    \centering
    \caption{Hyperparameters used for the experiments with the target complexity of a CCNN$_{6, 380}$.}
    \label{tab:hyperparams}
    \vspace{-2.5mm}
    \begin{small}
    \scalebox{0.8}{
    \begin{tabular}{lcccccccccc}
    \toprule
    \multirow{2}{*}{\sc{Dataset}} & \multirow{2}{*}{\sc{Epochs}} & \sc{Batch} & \sc{Learning} & \multirow{2}{*}{\sc{DropOut}} & \sc{Weight} & \multirow{2}{*}{$\omega_0$} & \multirow{2}{*}{$\lambda$} & \multirow{2}{*}{$\tau_\mathrm{resolution}$} & \multirow{2}{*}{$\tau_\mathrm{channel}$} & \multirow{2}{*}{$\tau_\mathrm{depth}$}\\
    & & \sc{Size} & \sc{Rate} & & \sc{Decay}\\
     \midrule
     \sc{CIFAR10} & 210 & 50 & 0.005 & 0.0 & 0.01 & 21.5{\sc{k}} & 0.1 & 50 & 50 & 16 \\
     \sc{CIFAR100} & 210 & 50 & 0.01 & 0.0 & 0.01 & 6.5{\sc{k}} & 0.1 & 25 & 25 & 8 \\
     \midrule
     \sc{DarcyFlow} & 310 & 12 & 0.01 & 0.2 & 0.0 & 7.5{\sc{k}} & 0.1 & 50 & 25 & 8\\
     \sc{Cosmic} & 310 & 4 & 0.01 & 0.2 & 0.01 & 5.5{\sc{k}} & 0.1 & 50 & 50 & 8\\
    \bottomrule
    \end{tabular}}
    \end{small}
    \vspace{-4mm}
\end{table*}

To overcome this issue, we limit the maximum size of the mask calculated by Eq.~\ref{eq:size_sigm} to be less or equal than the maximum allowable size, e.g., $\mathrm{size}(m){=}\min(\mathrm{size}(m), 280)$. It is important to note, however, that clipping the value of $\mathrm{size}$ directly would stop the gradient flow for parameter values leading to sizes larger 280. As a result, once the maximum size is reached, the mask would not be able to contract anymore. We avoid gradient flow stop by using clipping in combination with a straight-through estimator \citep{bengio2013estimating}. As a result, we are able to propagate the gradient across the clipping operation, and the resulting mask can still be modified even in the cropping operation is used.

\subsection{Hyperparameters and training configurations} \label{appx:hyperparams}
In this section, we include more information about the found hyperparameters, the values that were considered during hyperparameter tuning, and other training settings. The final hyperparameters used are listed in Table~\ref{tab:hyperparams}.

\textbf{Optimizer, learning rates and learning rate schedule.} All our models are optimized with AdamW \cite{loshchilov2017decoupled} in combination with a cosine annealing learning rate scheduler \cite{loshchilov2016sgdr}, and a linear learning rate warm-up stage of 10 epochs, except for {\tt ListOps}, {\tt Retrieval} and {\tt Path-X} for which we have a warm-up stage of 5 epochs.

\textbf{Regularization.} We utilize dropout \cite{srivastava2014dropout} --as shown in Fig.~\ref{fig:ccnn_architecture}-- as well as weight decay during training.

\subsubsection{Hyperparameter tuning}

\textbf{Frequency prior $\omega_0$.} The possible $\omega_0$ values explored in this work are $[1, 500, 1500, 2500, ... 28500, 29500, 30000]$.

\textbf{Tuning the value of $\lambda$.} $\lambda$ plays the role of controlling the weight of the computational loss $\gL_\mathrm{comp}$ relative to the task objective loss $\gL_\mathrm{obj}$. In this work, we find two settings which require different values of $\lambda$. One, given by the tasks that converge to a low prediction values relative to perfection, i.e., {\tt ListOps} and {\tt CIFAR100}, and for which the loss $\gL_\mathrm{obj}$ remains relatively high at the end of training. The other group is given by all the other tasks, which converge to high prediction values --many even obtaining a perfect accuracy on the train set--, and for which $\gL_\mathrm{obj}$ converges to values close to zero. For the first group, we require a higher value of $\lambda$ such that the computational complexity loss $\gL_\mathrm{comp}$ remains relevant to the optimization objective. The final values of $\lambda$ used are $5.0$ and $0.1$, respectively.

\textbf{Tuning the temperature of the Sigmoid masks $\tau$.} For the resolution mask, we consider three values of $\tau$, $[25, 50, 100]$ which correspond to a minimum size of $10\%$, $5\%$ and $2.5\%$ of the corresponding dimension. For the channel mask, we consider two values $\tau\in[25, 50]$ which correspond to a minimum size of $10\%$ and $5\%$ of the corresponding dimension, but observe early during tuning that models prefer $\tau{=}25$. For the depth dimension, which is much more sparse than the channel and resolution dimensions, we consider two values $\tau\in[8, 16]$, which result on a minimum depth of $2$ and $1$ layers, respectively. We observe early during tuning that models prefer $\tau{=}8$.

\textbf{Learning rate.} The possible learning rate values explored in this work are $[0.0001, 0.0005, 0.001, 0.005, 0.01, 0.02]$.

\textbf{Dropout.} The possible dropout values explored in this work are $[0.0, 0.1, 0.2, 0.3]$.

\textbf{Weight decay.} The possible weight decay values considered in this work are $[0.0, 0.0001, 0.001, 0.01, 0.05]$.

\end{document}